\documentclass[pdflatex,iicol,sn-mathphys-ay]{sn-jnl}
\usepackage{graphicx}
\usepackage{multirow}
\usepackage{amsmath,amssymb,amsfonts}
\usepackage{amsthm}
\usepackage{mathrsfs}
\usepackage[title]{appendix}
\usepackage{xcolor}
\usepackage{textcomp}
\usepackage{manyfoot}
\usepackage{booktabs}
\usepackage{algorithm}
\usepackage{listings}
\usepackage{bm}
\usepackage{colonequals}
\usepackage{algorithmic}
\usepackage{clrscode}

\theoremstyle{thmstyleone}%
\newtheorem{lemma}{Lemma}

\theoremstyle{thmstyletwo}

\newtheorem{remark}{Remark}
\newtheorem{corollary}{Corollary}

\theoremstyle{thmstylethree}
\newtheorem{definition}{Definition}

\def\pnmf{p_{\mathrm{PNMF}}}
\def\ptop{p_{\mathrm{MTM}}}
\def\L{{\bf L}}
\def\F{{\bf F}}

\def\X{{\bf X}}

\begin{document}

\title[Non-negative matrix factorization algorithms generally improve
  topic model fits]{Non-negative Matrix Factorization Algorithms Generally 
  Improve Topic Model Fits}


\author*[1]{\fnm{Peter} \sur{Carbonetto}}\email{pcarbo@uchicago.edu}

\author[1,2]{\fnm{Abhishek} \sur{Sarkar}}

\author[3]{\fnm{Zihao} \sur{Wang}}

\author[1,3]{\fnm{Matthew} \sur{Stephens}}

\affil*[1]{\orgdiv{Department of Human Genetics}, \orgname{University of Chicago}, \orgaddress{\city{Chicago}, \state{IL}, \country{USA}}}

\affil[2]{\orgname{Vesalius Therapeutics}, \orgaddress{\street{Street}, \city{Cambridge}, \state{MA}, \country{USA}}}

\affil[3]{\orgdiv{Department of Statistics}, \orgname{University of Chicago}, \orgaddress{\city{Chicago}, \state{IL}, \country{USA}}}


% Please provide an abstract of 150 to 250 words. The abstract should
% not contain any undefined abbreviations or unspecified references.
\abstract{\noindent In an effort to develop topic modeling methods that can be 
quickly applied to large data sets, we revisit the problem of
maximum-likelihood estimation in topic models. It is known, at least
informally, that maximum-likelihood estimation in topic models is
closely related to non-negative matrix factorization (NMF). Yet, to
our knowledge, this relationship has not been exploited previously to
fit topic models.
%
% NMF avoids the ``sum-to-one'' constraints on the topic model
% parameters, resulting in an optimization problem with simpler
% structure and more efficient computations.
%
We show that recent advances in NMF optimization methods can be
leveraged to fit topic models very efficiently, often resulting in
much better fits and in less time than existing algorithms for topic
models. We also formally make the connection between the NMF
optimization problem and maximum-likelihood estimation for the topic
model, and using this result we show that the expectation maximization
(EM) algorithm for the topic model is essentially the same as the
classic multiplicative updates for NMF.
%
% (the only difference being that the operations are performed in a
% different order).
%
Our methods are implemented in the R package ``fastTopics''.}

% Please provide 4 to 6 keywords which can be used for indexing purposes.

\maketitle

\section{Introduction}
\label{sec:intro}

The focus of this paper is the problem of computing maximum-likelihood
estimates (MLEs) of the parameters in a topic model given an $n \times
m$ matrix of counts.
%
% \citep{hofmann-1999}
%
Instead of directly computing the MLE for the topic model, we instead
propose to solve a similar problem with much simpler constraints on
the parameters: optimizing a non-negative matrix factorization (NMF)
based on a Poisson model of the data \citep{lee-seung-1999,
  lee-seung-2001, cichocki-2011, dhillon-2005, fevotte-2011,
  hien-2020}.
%
% This problem can be expressed as an optimization problem with linear
% constraints on the variables.
%
% (Note that, since finding the global maximum---that is, the
% MLE---is NP-hard, see \citealt{arora-2012}, we seek only to find a local
% maximum of the likelihood.)
%
% (This objective function can also be derived as a divergence
% measure \cite{cichocki-2011, dhillon-2005, fevotte-2011, hien-2020}.)
%
% We show that the Poisson NMF optimization problem is in fact
% equivalent to the maximum-likelihood estimation for the topic model.
%
Despite the fact that the topic model and Poisson NMF are well known
to be closely related \citep{buntine-2002, buntine-2006, canny-2004,
  ding-2008, faleiros-2016, gaussier-goutte-2005, gillis-book,
  zhou-2011}, as far as we are aware this relationship has not been
previously exploited for fitting topic models.
%
% We demonstrate the benefits of this approach in a variety of data
% sets, both real an simulated.
%

An intuition for the advantage of this approach is that the Poisson
NMF optimization problem lacks the ``sum-to-one'' constraints, which
complicate optimization. However, not all Poisson NMF algorithms
exploit this benefit. Indeed, the traditional way to solve Poisson
NMF---the ``multiplicative updates'' of \cite{lee-seung-2001}---is
equivalent to expectation maximization (EM) \citep{cemgil-2009}, and,
as we show, is closely related to the EM algorithms traditionally used
to fit topic models. (In fact, we show that these algorithms are
mostly the same except for the order in which the operations are
performed.)  Therefore, the Poisson NMF multiplicative updates are
expected to experience the same issues as EM, and this is indeed borne
out by our experiments. In contrast, other recently developed
algorithms for solving Poisson NMF based on co-ordinate descent (CD)
\citep{ccd, scd} do not have an existing counterpart for topic models.
%
% (Consider that co-ordinate descent cannot obviously deal
% with the sum-to-one constraints.)
%
It has been shown that CD algorithms can greatly outperform the
multiplicative updates for Poisson NMF \citep{hien-2020}. And here we
show that CD algorithms can also be leveraged to fit topic models very
efficiently, often resulting in much better fits and in less time than
the existing algorithms for topic models.

A maximum-likelihood approach to topic modeling is not new, of course;
one of the very first papers on topic modeling, \cite{hofmann-1999},
used a simple EM algorithm to obtain MLEs under the topic model. EM,
however, can be very slow to converge to a local maximum of the
likelihood \citep{redner-walker-1984, ma2000em, raz2020em,
  kunstner2021em, zhou-lange-2011, varadhan-2008, daarem}. The slow
convergence of EM is sometimes viewed as a feature, not a bug: ``early
stopping'' has been shown, both anecdotally and in theory, to result
in parameter estimates that better generalize to test sets---that is,
early stopping can {\em implicitly regularize} the MLEs
\citep{ali2019continuoustime, gunasekar2017implicit}. We show however
that this slow convergence can also sometimes cause the EM to get
``stuck'' in areas of the likelihood that are far away from a local
maximum, resulting in very poor parameter estimates. We also show that
the fast NMF algorithms can very quickly ``rescue'' the EM estimates,
resulting in parameter estimates that are very different from and much
better than the estimates produced by EM.

The maximum-likelihood approach we study in this paper contrasts with
the much more widely used variational inference approach for topic
models, i.e., latent Dirichlet allocation \citep{blei-2003, teh-2007,
  asuncion-2009}. The benefit of variational inference is that it
produces approximate posterior estimates of the model parameters,
which can help to address overfitting, stabilize parameter estimates,
and increase accuracy. However, the underlying computations for
variational inference are more complex, making the algorithms slower
and more challenging to apply to very large data sets. For these
reasons, ``online'' variational inference algorithms have been
developed
%
% that are much faster and can handle very large data sets
%
\citep{hoffman-2010, sato-2001}. But online learning algorithms bring
their own challenges; for example, unlike conventional optimization
algorithms, they do not guarantee that the objective will improve at
each iteration, and the results of online learning are often sensitive
to parameter tuning. (Markov chain Monte Carlo algorithms for
posterior inference in topic models have also been used in the past
e.g., \citealt{griffiths-2004}, but MCMC is typically more
computationally burdensome than variational inference.) Therefore, on
balance, maximum-likelihood estimation remains an attractive option
for many large data sets, especially when maximum-likelihood
estimation is implemented using fast NMF algorithms, as we show
here. Indeed, reframing the problem of fitting a topic model as an NMF
optimization problem has already enabled us and others to efficiently
fit topic models to very large single-cell data sets, in some cases
with $n, m \geq \mbox{100,000}$ \citep{chirichella2025integrated,
  dey-2017, gonzalez-blas-2019, gom_de, umans2025oxygen,
  meir2025early, gao2024dissection, popp2024celltype, liang2024eomes,
  zhao2024celltyperesolved, housman-2022, hung-2022, rhodes-2022,
  bastide2022tattooseq}.
%
% (See also \citealt{dey-2017,
% gonzalez-blas-2019} for earlier work on applying topic models to
% single-cell data.)
%
The numerical experiments in this paper include two single-cell data
sets, and demonstrate the benefits of applying fast NMF algorithms to
fit topic models for single-cell data sets.

% We propose to leverage these advances to fit topic models by taking
% the following simple approach: first, we fit a Poisson NMF model by
% solving \eqref{eq:poisson-nmf-problem}; next, we recover the
% equivalent topic model by a simple transformation (see
% Definition~\ref{def:poisson2multinomial}). The benefit of this
% approach is that the Poisson NMF optimization problem has simpler
% constraints and a structure that can be exploited for more efficient
% computation.
%
The algorithms for fitting topic models and Poisson NMF described in
this paper are implemented in an R package, {\tt fastTopics},
which available on CRAN
(\url{https://cran.r-project.org/package=fastTopics}) and on GitHub
(\url{https://github.com/stephenslab/fastTopics/}).

% Note that section titles should capitalize each word.
\section{Poisson NMF and the multinomial topic model}
\label{sec:models}

% Although the topic model was originally proposed with the aim of
% discovering patterns from collections of text documents, the topic
% model can also be viewed more generally as a model for learning a
% reduced representation of count data in which each document is
% represented by a linear combination of ``topics''
% \citep{hofmann-1999}.

In the following, we provide side-by-side descriptions of the topic
model and Poisson NMF to highlight their close connection. While
formal and informal connections between these two models have been
made previously \citep{buntine-2002, buntine-2006, canny-2004,
  ding-2008, faleiros-2016, gaussier-goutte-2005, zhou-2011}.
%
% these previous papers draw connections between the algorithms and/or
% stationary points of the objective functions.
%
We provide a simple and more general result relating the likelihoods
of the two models (Lemma~\ref{lemma:pnmf-plsi-equivalence}), which we
view as a more fundamental result underlying previous
results.\footnote{Recent papers have also studied the problem of
identifying ``anchor words,'' which are words that appear in exactly
one topic. In this setting, there is also a close relationship between
the algorithms for identifying anchor words and the algorithms for
identifying ``separable'' non-negative factors \citep{arora-2012,
  arora-2013, donoho-stodden-2003, gillis-book, gillis-2015}.}

Let $\X \in {\bf R}_{+}^{n \times m}$ denote an $n \times m$ matrix of
observed counts $x_{ij}$.  For example, when analyzing text documents,
$n$ is the number of documents, $m$ is the number of unique terms, and
$x_{ij}$ is the number of times term $j$ occurs in document $i$. Both
Poisson NMF and the topic model can be seen as fitting different---but
closely related---models of $\X$.

% in which entry $(i,j)$ records the number of occurrences of term $j$
% in document $i$

The Poisson NMF model has parameters that are non-negative matrices,
${\bf H} \in {\bf R}_{+}^{n \times K}$ and ${\bf W} \in {\bf R}_{+}^{m
  \times K}$, where ${\bf R}_{+}^{r \times c}$ denotes the set of
non-negative, real matrices with $r$ rows and $c$ columns. Given a $K
\geq 1$, the Poisson NMF model is
\begin{equation}
\begin{aligned}
x_{ij} &\mid {\bf H}, {\bf W} \sim
\mathrm{Pois}(\lambda_{ij}) \\
\lambda_{ij} &= ({\bf HW}^T)_{ij} = \sum_{k=1}^K h_{ik} w_{jk},
\end{aligned}
\label{eq:poisson-nmf}
\end{equation}
where $h_{ij}, w_{jk}$ denote elements of matrices ${\bf H}, {\bf W}$,
and $\mathrm{Pois}(\lambda)$ denotes the Poisson distribution with
rate $\lambda$. Poisson NMF can be viewed as a rank-$K$ matrix
factorization by noting that \eqref{eq:poisson-nmf} implies $E[\X] =
{\bf HW}^T$. So fitting a Poisson NMF essentially seeks values of
${\bf H}$ and ${\bf W}$ such that $\X \approx {\bf HW}^T$.\footnote{In
descriptions of NMF, it is more common to represent data vectors
(e.g., documents) as {\em columns} of ${\bf X}$ (e.g.,
\citealt{gillis-book, kim-2014, lee-seung-2001}), in which case one
would write ${\bf X} \approx {\bf W} {\bf H}^T$. Here, we represent
documents as {\em rows} of ${\bf X}$, following 
\cite{hofmann-1999, taddy-2012}. Due to the symmetry of Poisson NMF
\eqref{eq:poisson-nmf}, it makes no difference if we fit ${\bf X}
\approx {\bf H} {\bf W}^T$ or ${\bf X}^T \approx {\bf H} {\bf
  W}^T$. It only matters when connecting Poisson NMF to the topic
model.} Computing an MLE for the Poisson NMF model reduces to the
following {\em bound-constrained} optimization problem:
\begin{equation}
\begin{array}{ll}
\mbox{minimize} & \ell({\bf X}; {\bf H}, {\bf W}) \\
\mbox{subject to} & {\bf H} \geq {\bm 0}, {\bf W} \geq {\bm 0},
\end{array}
\label{eq:poisson-nmf-problem}
\end{equation}
in which the objective function is
\begin{align}
\ell({\bf X}; {\bf H}, {\bf W}) \colonequals&\;
\phi({\bf X}; {\bf H}, {\bf W}) + 
\|{\bf H}{\bf W}^T \|_{1,1}
\label{eq:poisson-nmf-objective} \\
\phi({\bf X}; {\bf H}, {\bf W}) \colonequals&\;
-\sum_{i=1}^n \sum_{j=1}^m x_{ij} \log {\bm h}_i^T {\bm w}_j,
\label{eq:phi}
\end{align}
where $\|{\bf A}\|_{1,1} = \sum_{i = 1}^n \sum_{j = 1}^m |a_{ij}|$ is
the $L_{1,1}$ norm of $n \times m$ matrix ${\bf A}$, and
 ${\bm h}_i, {\bm w}_j$ denote, respectively, the $i$th row of
${\bf H}$ and the $j$th row of ${\bf W}$.

%
% Topic models analyze a term-document count matrix to learn a
% representation of each document as a non-negative, linear combination
% of ``topics'' \citep{buntine-2002, griffiths-2004, minka-2002}.
%
Like Poisson NMF, the multinomial topic model is also
parameterized by two non-negative matrices, $\L \in {\bf
R}_{+}^{n \times K}$, $\F \in {\bf R}_{+}^{m \times K}$, but the
elements of these two matrices must satisfy additional ``sum-to-one''
constraints:
\begin{equation}
\label{eq:sum_to_one}
\sum_{j=1}^m f_{jk} = 1,
\quad \sum_{k=1}^K l_{ik} = 1. 
\end{equation} 
Most variations of the topic model, including the aspect model
\citep{hofmann-puzicha-jordan-1999}, probabilistic latent semantic
indexing \citep{hofmann-2001, hoffman-2010, hofmann-1999} and latent
Dirichlet allocation \citep{blei-2003}, are based on the same basic
model: a multinomial distribution of the counts. We therefore refer to
this model as the {\em multinomial topic model.}
%
% While there have been many different approaches to topic modeling,
% with different fitting procedures and different prior
% distributions---
%
Given a $K \geq 2$, the multinomial topic model is
\begin{equation}
\begin{aligned}
x_{i1}, \ldots, x_{im} &\mid {\bf L}, {\bf F} \sim
\mathrm{Multin}(t_i; \pi_{i1}, \ldots, \pi_{im}) \\
\pi_{ij} &= (\L \F^T)_{ij} =
\sum_{k=1}^K l_{ik} f_{jk},
\end{aligned}
\label{eq:multinomial-topic-model}
\end{equation}
in which $\mathrm{Multin}(n; \pi_1, \ldots, \pi_m)$ is the multinomal
distribution with sample size $n$ and probabilities $\pi_1, \ldots,
\pi_m$, and $t_i \colonequals \sum_{j=1}^m x_{ij}$. The multinomial
topic model is also a matrix factorization because we have that ${\bf
  \Pi} = \L \F^T$, where ${\bf \Pi}$ denotes the matrix of multinomial
probabilities $\pi_{ij}$ \citep{hofmann-1999, singh-2008,
  steyvers-2006}. Computing an MLE for the multinomial topic model
reduces to a {\em linearly constrained} optimization problem,
\begin{equation}
\begin{array}{ll}
\mbox{minimize} & \phi({\bf X}; {\bf L}, {\bf F}) \\
\mbox{subject to} & {\bf L} \bm{1}_K = \bm{1}_n \\
& {\bf F}^T \bm{1}_m = \bm{1}_K \\
& {\bf L} \geq \bm{0}, {\bf F} \geq \bm{0}, 
\end{array}
\label{eq:main-problem}
\end{equation}
in which $\bm{1}_d = (1, \ldots, 1)^T$ denotes a column vector of ones
of length $d$, and $\phi$ was defined in \eqref{eq:phi}.

% 
% \begin{equation}
% \phi({\bf L}, {\bf F}) = 
% -\sum_{i=1}^n \sum_{j=1}^m x_{ij} \log {\bm l}_i^T {\bm f}_j,
% \end{equation}
% where ${\bm l}_i, {\bm f}_j$ denote, respectively, the $i$th row of
% ${\bf L}$ and the $j$th row of ${\bf F}$.
% 
Now we connect the Poisson non-negative matrix factorization
%
% with parameters ${\bf H}, {\bf W}$
%
to the multinomial topic model matrix factorization.
%
% with parameters ${\bf L}, {\bf F}$
%
To do so, we define a mapping between the parameter spaces for the two
models (Definition~\ref{def:poisson2multinomial}), and then we state an
equivalence between their likelihoods (Lemma
\ref{lemma:pnmf-plsi-equivalence}), which leads to an equivalence in
their MLEs (Corollary \ref{cor:mle}).

\begin{definition}[Poisson NMF to multinomial topic model reparameterization]
\label{def:poisson2multinomial}
\rm Let ${\bf R}_{++}^d$ denote the set of positive real vectors of
length $d$, let ${\bf R}_{\rm row}^{r \times c}$ denote the set of $r
\times c$ row-normalized matrices (non-negative matrices ${\bf A}$
with the property that the elements in each row of ${\bf A}$ sum to
1), and let ${\bf R}_{\rm col}^{r \times c}$ denote the set of $r
\times c$ column-normalized matrices (non-negative
matrices ${\bf A}$ with the property that the elements in each column
of ${\bf A}$ sum to 1). For $K \geq 2$, ${\bf H} \in {\bf
R}_{+}^{n \times K}$, ${\bf W} \in {\bf R}_{+}^{m \times K}$, define 
mapping $\proc{PNMF-to-MTM} : {\bf H}, {\bf W} \mapsto \L, \F, {\bm
s}, {\bm u}$, with $\L \in {\bf R}_{\mathrm{row}}^{n \times K}$, ${\bf
F} \in {\bf R}_{\mathrm{col}}^{m \times K}$, ${\bm s} \in {\bf
R}_{++}^n$, ${\bm u}
\in {\bf R}_{++}^K$ by the following procedure:
\begin{codebox}
\Procname{$\proc{PNMF-to-MTM}({\bf H}, {\bf W})$}
\li ${\bf F}, {\bm u} \leftarrow \proc{Normalize-Cols}({\bf W})$
\li ${\bf U} \leftarrow \mathrm{diag}({\bm u})$
\li ${\bf L}, {\bm s} \leftarrow \proc{Normalize-Rows}({\bf H} {\bf U})$
\li \Return $(\L, \F, {\bm s}, {\bm u})$
\end{codebox}
This procedure has two subroutines, defined as follows:
  $\proc{Normalize-Rows}({\bf A})$ returns a vector ${\bm y} \in {\bf
  R}_{++}^r$ containing the row sums of $r \times c$ matrix ${\bf A}$,
  $y_i = \sum_{j=1}^c a_{ij}$, and ${\bf B} \in {\bf R}_{\rm
  row}^{r \times c}$, a row-normalized matrix with entries $b_{ij} =
  a_{ij} / y_i$; and $\proc{Normalize-Cols}({\bf A})$ returns a vector
  ${\bm y} \in {\bf R}_{++}^c$ containing the column sums of ${\bf
  A}$, $y_j =
\sum_{i=1}^r a_{ij}$, and ${\bf B} \in {\bf R}_{\rm col}^{r \times
  c}$, a column-normalized matrix with entries $b_{ij} = a_{ij} /
y_j$. We also define $\mathrm{diag}({\bm a})$ as the $n \times n$
diagonal matrix ${\bf A}$ with diagonal entries given by the elements
of vector ${\bm a}$.
\end{definition}
 $\proc{Normalize-Rows}$ also defines a mapping $\proc{Normalize-Rows}
: {\bf A} \mapsto {\bf B}, {\bm y}$, with ${\bf A} \in {\bf
R}_{+}^{r \times c}$, ${\bf B} \in {\bf R}_{\mathrm{row}}^{r \times
c}$, ${\bf y} \in {\bf R}_{++}^r$. If each row of ${\bf A}$ has at
least one positive element, then this mapping is one-to-one, and
therefore $\proc{Normalize-Rows}$ defines a {\em change of variables}
from non-negative matrices ${\bf A}$ to row-normalized matrices ${\bf
B}$ and positive vectors ${\bm y}$. Similarly, $\proc{Normalize-Cols}
: {\bf A} \mapsto {\bf B}, {\bm y}$ defines a change of variables from
non-negative matrices ${\bf A}$ to column-normalized matrices ${\bf
B}$ and positive vectors ${\bm y}$ (provided that each column of ${\bf
A}$ has at least one positive element). These together imply that
$\proc{MTM-to-PNMF}$ defines a change of variables from 
non-negative matrices ${\bf H}, {\bf W}$ to positive vectors ${\bm
s}, {\bm u}$, row-normalized matrices $\L$, and column-normalized
matrices $\F$. The ${\bf F}$ and ${\bf L}$ satisfy the sum-to-one
constraints \eqref{eq:sum_to_one}.
%
% so long as we impose the additional restriction that each row of
% ${\bf H}$ has at least one positive entry and each column of ${\bf
% W}$ has at least one positive entry.
%
The inverse mapping, $\proc{MTM-to-PNMF} \colonequals
\proc{PNMF-to-MTM}^{-1} : \L, \F, {\bm s}, {\bm u} \mapsto {\bf H},
     {\bf W}$, is ${\bf W} \leftarrow \F {\bf U}$, ${\bf H} \leftarrow
     {\bf S} \L {\bf U}^{-1}$, where ${\bf U} \colonequals
     \mathrm{diag}({\bm u})$, ${\bf S} \colonequals \mathrm{diag}({\bm
       s})$.

\begin{lemma}[Equivalence of Poisson NMF and multinomial topic model 
  likelihoods]
\label{lemma:pnmf-plsi-equivalence}
\rm Denote the Poisson NMF likelihood by $\pnmf({\X} \mid {\bf H},
    {\bf W})$ and denote the multinomial topic model likelihood by
    $\ptop({\X} \mid \L, \F)$. Assume ${\bf H} \in {\bf R}_{+}^{n
      \times K}$ and ${\bf W} \in {\bf R}_{+}^{m \times K}$, define
    $t_i \colonequals \sum_{j=1}^m x_{ij}$, and let $\L, \F, {\bm s},
    {\bm u}$ be the result of applying $\proc{PNMF-to-MTM}$ to ${\bf
      H}, {\bf W}$. Then we have that
\begin{equation}
\pnmf({\X} \mid {\bf H}, {\bf W}) =
\ptop({\X} \mid \L, \F) \prod_{i=1}^n \mathrm{Pois}(t_i; s_i),
\label{eq:pnmf-plsi-equivalence}
\end{equation}
where $\mathrm{Pois}(x; \lambda)$ denotes the probability mass
function of the Poisson distribution at $x$ with rate $\lambda$.
\end{lemma}
\begin{proof}
The result is obtained by applying the following identity relating the
multinomial and Poisson distributions \citep{fisher-1922, good-1986}:
\begin{equation}
\prod_{j=1}^m
\mathrm{Pois}(x_j; \lambda_j) = 
\mathrm{Multin}(\bm{x}; t, \lambda_1/s, \ldots, \lambda_m/s)
\mathrm{Pois}(t; s),
\label{eq:multinom-pois-equivalence}
\end{equation}
where $\bm{x} = (x_1, \dots, x_m)$, $\lambda_1, \ldots, \lambda_m \in
{\bf R}_{+}$, $s \colonequals \sum_{j=1}^m \lambda_j$, $t \colonequals
\sum_{j=1}^m x_j$, and $\mathrm{Multin}({\bm x}; n, \pi_1, \ldots,
\pi_m)$ denotes the probability mass function of the multinomial
distribution at ${\bm x} = (x_1, \ldots, x_m)$ with sample size $n$ and
probabilities $\pi_1, \ldots, \pi_m$.
\end{proof}

% Note that for computational reasons the multinomial distribution is
% sometimes approximated by a product of Poissons
% (e.g., \citealt{townes-2019}), but there is no need for this
% approximation here.

Now we use this lemma to justify solving the Poisson NMF optimization
problem in order to achieve maximum-likelihood estimation in the
multinomial topic model.  First, consider an {\em augmented form} of the
multinomial topic model optimization problem:
\begin{equation}
\begin{array}{ll}
\mbox{minimize} & \phi_{\mathrm{aug}}({\bf X}; {\bf L}, {\bf F}, {\bm s}) \\
\mbox{subject to} & {\bf L} \bm{1}_K = \bm{1}_n \\
& {\bf F}^T \bm{1}_m = \bm{1}_K \\
& {\bf L} \geq \bm{0}, {\bf F} \geq \bm{0}, {\bm s} \geq \bm{0}, 
\end{array}
\label{eq:augmented-problem}
\end{equation}
in which the augmented objective is
\begin{align}
\phi_{\mathrm{aug}}({\bf X}; {\bf L}, {\bf F}, {\bm s}) \colonequals 
\phi({\bf X}; {\bf L}, {\bf F}) + \psi({\bf X}; {\bm s}) 
\label{eq:augmented-objective} \\
\psi({\bf X}; {\bm s}) \colonequals
\sum_{i=1}^n s_i - \sum_{i=1}^n \sum_{j=1}^m x_{ij} \log s_i.
\label{eq:psi}
\end{align}
Notice that solutions to \eqref{eq:augmented-problem} are also
solutions to \eqref{eq:main-problem} because the objective and
constraints for the ${\bf L}$ and ${\bf F}$ have not changed. The
Poisson NMF objective $\ell({\bf X}; {\bf H}, {\bf W})$ is equal to
the Poisson log-likelihood, $\log \pnmf({\X} \mid {\bf H}, {\bf W})$
(ignoring constant terms), and the multinomial topic model augmented
objective $\phi_{\mathrm{aug}}({\bf X}; \L, \F, {\bf s})$ is equal
to the logarithm of the right-hand side of
\eqref{eq:pnmf-plsi-equivalence} (again, ignoring constant
terms). This means that any optimization algorithm that improves the
Poisson NMF objective $\ell({\bf X}; {\bf H}, {\bf W})$ will also
improve the augmented objective $\phi_{\mathrm{aug}}({\bf X}; \L, \F,
{\bf s})$ so long as $\proc{PNMF-to-MTM}$ is used to to recover $\L,
\F, {\bm s}$ from ${\bf H}, {\bf W}$. We formalize the relationship
between the two optimization problems in the following corollary.

\begin{corollary}[Relationship between MLEs for Poisson NMF and  
multinomial topic model] \label{cor:mle} \rm Let $\hat{\bf H} \in {\bf
    R}_{+}^{n \times K}, \hat{\bf W} \in {\bf R}_{+}^{m \times K}$
  denote MLEs for the Poisson NMF model,\footnote{The notation
    $\hat{\theta} \in \mathrm{argmax}_{\theta} \, f(\theta)$ means
    $f(\hat{\theta}) \geq f(\theta)$ for all $\theta$, and accounts
    for the fact that an MLE may not be unique due to
    non-identifiability.}
\begin{equation}
\hat{\bf H}, \hat{\bf W} \in
\underset{{\bf H} \in {\bf R}_{+}^{n \times K}, 
          {\bf W} \in {\bf R}_{+}^{m \times K}}
         {\mathrm{argmax}} 
\pnmf(\X \mid {\bf H}, {\bf W}).
\label{eq:mle-pnmf}
\end{equation}
Equivalently, $\hat{\bf H}, \hat{\bf W}$ can be defined as a solution
to \eqref{eq:poisson-nmf-problem}. If $\hat{\L}, \hat{\F}$ are
obtained by applying $\proc{PNMF-to-MTM}$ to $\hat{\bf H}, \hat{\bf
  W}$, then these are also MLEs for the multinomial topic model,
\begin{equation}
\hat{\L}, \hat{\F} \in
\underset{{\bf L} \,\in\, {\bf R}_{\mathrm{row}}^{n \times K}, \,
          {\bf F} \,\in\, {\bf R}_{\mathrm{col}}^{m \times K}}
         {\mathrm{argmax}} \;
\ptop(\X \mid \L, \F).
\label{eq:mle-multinomial-topic-model}
\end{equation}
Equivalently, $\hat{\L}, \hat{\F}$ are a solution to
\eqref{eq:main-problem}.

Conversely, let $\hat{\L} \in {\bf R}_{\mathrm{row}}^{n \times K},
\hat{\F} \in {\bf R}_{\mathrm{col}}^{m \times K}$ denote multinomial
topic model MLEs, set $\hat{\bm s} = {\bm t} \colonequals (t_1,
\ldots, t_n)$, and choose any $\hat{\bm u} \in {\bf R}_{++}^K$. If
$\hat{\bf H}, \hat{\bf W}$ are obtained by applying
$\proc{MTM-to-PNMF}$ to $\hat{\L}, \hat{\F}, \hat{\bm s}, \hat{\bm
  u}$, these are also Poisson NMF MLEs \eqref{eq:mle-pnmf}.
\end{corollary}

\begin{proof}
We prove this result using ``equivalent optimization problems''
\citep{boyd}. Since $\proc{PNMF-to-MTM}$ defines a change of
variables, we can apply the change of variables to
\eqref{eq:augmented-problem} to obtain an equivalent optimization problem 
with optimization variables ${\bf H}, {\bf W}$:
\begin{equation}
\begin{array}{ll}
\mbox{minimize} & 
\phi_{\mathrm{aug}}({\bf X}; \proc{PNMF-to-MTM}({\bf H}, {\bf W})) \\
\mbox{subject to} & {\bf H} \geq {\bm 0}, {\bf W} \geq {\bm 0},
\end{array}
\label{eq:equivalent-optimization-problem-1}
\end{equation}
in which the ${\bm u}$ returned by \proc{PNMF-to-MTM} is ignored.
% (Since the sum-to-one constraints are satisfied by construction, only
% the non-negativity constraints are needed.)
From Lemma \ref{lemma:pnmf-plsi-equivalence}, we can rewrite
\eqref{eq:equivalent-optimization-problem-1} as
\begin{equation}
\begin{array}{ll}
\mbox{minimize} & \ell({\bf X}; {\bf H}, {\bf W}) + \mbox{const} \\
\mbox{subject to} & {\bf H} \geq {\bm 0}, {\bf W} \geq {\bm 0},
\end{array}
\label{eq:equivalent-optimization-problem-2}
\end{equation}
which is exactly the Poisson NMF optimization problem (ignoring
terms that do not depend on ${\bf H}$ or ${\bf W}$). Therefore, the
augmented optimization problem
\eqref{eq:augmented-problem} and the Poisson NMF optimization problem
\eqref{eq:poisson-nmf-problem} are related to each other by the change
of variables $\proc{PNMF-to-MTM}({\bf H}, {\bf W}) = (\L, \F, {\bm s},
{\bm u})$. Since solutions to the augmented optimization problem
\eqref{eq:augmented-problem} are also solutions to the original
problem \eqref{eq:main-problem}, it follows that Poisson NMF MLEs
$\hat{\bf H}, \hat{\bf W}$ recover multinomial topic model MLEs
$\hat{\L}, \hat{\F}$. The reverse---that multinomial topic model MLEs
$\hat{\L}, \hat{\F}$ recover Poisson NMF MLEs $\hat{\bf H}, \hat{\bf
  W}$---requires the additional step of solving $\hat{\bf s}
\colonequals \mathrm{argmin}_{{\bm s} \,\in\, {\bf R}_{++}^n}
\psi({\bm s}) = \mathrm{argmax}_{{\bm s} \,\in\, {\bf R}_{++}^n}
\prod_{i=1}^n \mathrm{Pois}(t_i; s_i)$, which has unique solution
$\hat{\bm s} = {\bm t}$ provided that $t_1, \ldots, t_n > 0$.
\end{proof}

\begin{remark}
\rm Since ${\bf H}, {\bf W}$ are not uniquely identifiable---consider
that multiplying the $k$th column of ${\bf H}$ by $a_k \neq 0$ and
dividing the $k$th column of ${\bf W}$ by $a_k$ does not change ${\bf
  H} {\bf W}^T$---one way to avoid this non-identifiability is to
impose constraints or penalty terms to the objective. However,
introducing constraints or penalties on ${\bf H}, {\bf W}$ (or $\L,
\F$) may break the above equivalence. We note one form of penalized
objective that preserves the equivalence:
\begin{equation}
\phi^{\star}({\bf X}; \L, \F) \colonequals \phi({\bf X}; \L, \F) 
+ \rho^{\mathrm{MTM}}({\bf F}),
\label{eq:loss-mtm-penalized}
\end{equation}
where
\begin{equation}
\rho^{\mathrm{MTM}}({\bf F}) \colonequals
-\sum_{j=1}^m \sum_{k=1}^K (a_{jk} - 1) \log f_{jk},
\end{equation}
and $a_{jk} > 1$, $j = 1, \ldots, m$, $k = 1, \ldots, K$. The
equivalent penalized objective for Poisson NMF is
\begin{equation}
\ell^{\star}({\bf X}; {\bf H}, {\bf W}) \colonequals 
\ell({\bf X}; {\bf H}, {\bf W}) + \rho^{\mathrm{PNMF}}({\bf W}),
\label{eq:loss-pnmf-penalized}
\end{equation}
where
\begin{align}
\rho^{\mathrm{PNMF}}({\bf W}) \colonequals\;&
- \sum_{j=1}^m \sum_{k=1}^K (a_{jk} - 1) \log w_{jk} 
\nonumber \\
& + \sum_{j=1}^m \sum_{k=1}^K b_k w_{jk},
\end{align}
and $b_k > 0$, $k = 1, \ldots, K$. The $a_{jk}, b_k$ control the shape
and strength of these penalties. (Setting $a_{jk} = 1, b_k = 0$
recovers the unpenalized objectives.) Minimizing $\phi^{\star}({\bf
  X}; \L, \F)$ corresponds to MAP estimation of $\L, \F$ with
Dirichlet priors on $\F$ and uniform priors on $\L$
\citep{sontag-roy-2011}, and minimizing $\ell^{\star}({\bf X}; {\bf
  H}, {\bf W})$ corresponds to MAP estimation of ${\bf W}, {\bf H}$
with gamma priors on ${\bf W}$ and uniform priors on ${\bf H}$
\citep{canny-2004, cemgil-2009, ma-2011}. The equivalence of MAP
estimation with these specific priors i is a slight generalization of
Corollary \ref{cor:mle} (see Appendix \ref{sec:derivations}).
\end{remark}

Lemma \ref{lemma:pnmf-plsi-equivalence} and Corollary \ref{cor:mle}
are more general than previous results \citep{buntine-2006, ding-2008,
  gaussier-goutte-2005, gillis-book} because they apply to {\em any}
${\bf H}, {\bf W}$ and ${\bf L}, {\bf F}$ from Definition
\ref{def:poisson2multinomial}, not only a fixed point of the
likelihood or objective. See \cite{buntine-2002, faleiros-2016,
  zhou-2011} for other related results.

In short, Lemma \ref{lemma:pnmf-plsi-equivalence} tells us that
Poisson NMF and the multinomial topic model are Poisson and
multinomial formulations of the same matrix factorization method. In
particular, their shared ability to recover a decomposition into
``parts'' or ``topics'' is suggested by these formal connections.

Although Poisson NMF and the multinomial topic model achieve similar
ends,
% 
% and are identical for maximum-likelihood estimation and some
% forms of MAP estimation
%
the two methods still possess different advantages: Poisson NMF has an
advantage in computation because it avoids the sum-to-one constraints,
whereas the multinomial topic model has the advantage in
interpretation because the parameters $f_{jk}, l_{ik}$ can be compared
across topics $k$ whereas the Poisson NMF parameters $h_{ik}, w_{jk}$
cannot due to the undetermined column-scaling ${\bm u}$. Therefore, by
switching between the two models, we can have the advantages of both.
%
% Specifically, $f_{jk}$ is the frequency of word $j$ in topic $k$,
% which can be compared to $f_{jk'}$ in other topics $k' \neq k$, and
% $l_{ik}$ is the proportion of document $i$ attributed to topic $k$,
% which can be compared to $l_{i'k}$ in different documents $i' \neq i$.
%

\section{Poisson NMF algorithms}
\label{sec:algorithms}

Corollary \ref{cor:mle} implies that {\em any algorithm for
  maximum-likelihood estimation in Poisson NMF is also an algorithm
  for maximum-likelihood estimation in the multinomial topic model.}
  (This also means that the NP-hardness \citealt{arora-2012,
  vavasis-2010} of the two problems is related.) Fitting the Poisson
  NMF model involves solving \eqref{eq:poisson-nmf-problem}, which we
  restate here in a slightly different way:
\begin{equation} 
\begin{array}{ll}
\mbox{minimize} &
\displaystyle \ell({\bf X}; {\bf H},{\bf W}) =
\sum_{i=1}^n \sum_{j=1}^m {\bm h}_i^T {\bm w}_j 
- x_{ij} \log ({\bm h}_i^T {\bm w}_j) \\
\mbox{subject to} & {\bf H} \geq {\bf 0}, {\bf W} \geq {\bf 0},
\end{array}
\label{eq:problem}
\end{equation}
Here, ${\bm h}_i$ and ${\bm w}_j$ denote column vectors containing,
respectively, the $i$th row of ${\bf H}$ and the $j$th row of ${\bf
W}$, and we assume $K \geq 2$.

% The loss function $\ell({\bf H}, {\bf W})$ is equal to
% $-\log p_{\mathrm{PNMF}}({\bf X} \mid {\bf H}, {\bf W})$ after
% removing terms that do not depend on ${\bf H}$ or ${\bf W}$.

To facilitate comparisons of different algorithms for solving
\eqref{eq:problem}, in the next section we introduce an ``Alternating
Poisson Regression'' framework for solving \eqref{eq:problem}, then we
describe the algorithms we have implemented, drawing on recent work
% \citep{ang-2019, hien-2020, ccd,
%   scd}
and our own experimentation. See also \cite{hien-2020} for a detailed
comparison of Poisson NMF algorithms.

\subsection{Alternating Poisson Regression for Poisson NMF}
\label{sec:bcd}

Alternating Poisson Regression arises from solving \eqref{eq:problem}
by switching between optimizing over ${\bf H}$ with ${\bf W}$ fixed,
and optimizing over ${\bf W}$ with ${\bf H}$ fixed. This is an example
of a block-coordinate descent algorithm \citep{wright-2015,
bertsekas-1999} 
%
% (also ``nonlinear Gauss-Seidel''), 
%
where the two ``blocks'' are ${\bf H}$ and ${\bf W}$. It is analogous
to ``alternating least squares'' for matrix factorization with
Gaussian errors.

We point out two simple but important facts. First, by symmetry of
\eqref{eq:problem}, optimizing ${\bf H}$ given ${\bf W}$ has the same
form as optimizing ${\bf W}$ given ${\bf H}$.
%
% which simplifies implementation. 
%
Second, because of the separability of the sum in
\eqref{eq:problem}, optimizing ${\bf W}$ given ${\bf H}$ splits into
$m$ independent $K$-dimensional subproblems of the following form:
\begin{equation} 
\begin{array}{l@{\;}l}
\mbox{minimize} & \ell_j({\bm w}_j) \colonequals
\sum_{i=1}^n {\bm h}_i^T{\bm w}_j - x_{ij} \log ({\bm h}_i^T{\bm w}_j) \\
\mbox{subject to} & {\bm w}_j \geq {\bm 0},
\end{array}
\label{eq:subproblem}
\end{equation}
for $j = 1, \dots, m$. (And similarly for optimizing ${\bf H}$ given
${\bf W}$.) Because the $m$ subproblems \eqref{eq:subproblem} are
independent, their solutions can be pursued in parallel. While both of
these observations are simple, neither of them hold for the
multinomial topic model due to the sum-to-one constraints.

Subproblem \eqref{eq:subproblem} is itself a well-studied
maximum-likelihood estimation problem \citep{em-book,
lange-carson-1984, lucy-1974, molina-2001, richardson-1972,
shepp-vardi-1982, vardi-1985}, and it is equivalent to computing an MLE of
${\bm b} \colonequals (b_1, \ldots, b_K)^T \geq {\bm 0}$ in an
additive Poisson regression model:
\begin{equation}
\begin{aligned}
y_i   &\sim \mathrm{Pois}(\mu_i) \\
\mu_i &= \textstyle \sum_{k=1}^K a_{ik} b_k,
\end{aligned}
\label{eq:apr}
\end{equation}
in which ${\bm y} = (y_1, \ldots, y_n)^T \in {\bf R}_{+}^n$ and ${\bf
A}
\in {\bf R}_{+}^{n \times K}$. 
%
% (We say ``additive'' to distinguish the model from the more common
% multiplicative formulation; e.g., \citealt{glm}.)
%
Consider a function that returns an MLE of ${\bm b}$ for \eqref{eq:apr},
\begin{equation}
\proc{Fit-Pois-Reg}({\bf A}, {\bm y})
\colonequals \underset{{\bm b} \, \in \, {\bf R}_{+}^K}{\mathrm{argmax}} \;
p_{\mathrm{PR}}({\bm y} \mid
{\bf A}, {\bm b}),
\label{eq:fit-pois-reg}
\end{equation} 
where $p_{\mathrm{PR}}({\bm y} \mid {\bf A}, {\bm b})$ denotes the
likelihood under the Poisson regression model \eqref{eq:apr}. Any
algorithm that solves \eqref{eq:fit-pois-reg} be can applied
iteratively to solve the Poisson NMF problem \eqref{eq:problem}. This
idea, which we call ``alternating Poisson regression for Poisson
NMF'', is formalized in Algorithm \ref{alg:bcd}.
%
% (Similarly, Frobenius-norm NMF is often solved
% by iteratively solving a series of non-negative least squares
% problems; see \citealt{gillis-2014, kim-2014}.)
%

\begin{algorithm}[t]
\caption{Alternating Poisson Regression for Poisson NMF. Here, ${\bm
    x}_i$ denotes a row of ${\X}$ and ${\bm x}_j$ denotes a column of ${\bf
    X}$.}
\label{alg:bcd}
\begin{algorithmic}
\raggedright
\REQUIRE ${\X} \in {\bf R}_{+}^{n \times m}$, initial estimates
${\bf H}^{(0)} \in {\bf R}_{+}^{n \times K}$, ${\bf W}^{(0)} \in {\bf
  R}_{+}^{m \times K}$, 
and a function $\proc{Fit-Pois-Reg}({\bf A}, {\bm y})$ that 
returns an MLE of ${\bm b}$ in \eqref{eq:apr}.
\FOR{$t = 1, 2, \ldots$}
% \FOR[can be performed in parallel]{$i = 1, \ldots, n$}
\FOR{$i = 1, \ldots, n$} 
\STATE ${\bm h}_i \leftarrow \proc{Fit-Pois-Reg}({\bf W}^{(t-1)}, {\bm x}_i)$
\STATE Store ${\bm h}_i$ in $i$th row of ${\bf H}^{(t)}$
\ENDFOR
% \FOR[can be performed in parallel]{$j = 1, \ldots, m$}
\FOR{$j = 1, \ldots, m$} 
\STATE ${\bm w}_j \leftarrow \proc{Fit-Pois-Reg}({\bf H}^{(t)}, {\bm x}_j)$
\STATE Store ${\bm w}_j$ in $j$th row of ${\bf W}^{(t)}$
\ENDFOR 
\ENDFOR
\RETURN ${\bf H}^{(t)}$, ${\bf W}^{(t)}$
\end{algorithmic}
\end{algorithm}

\subsection{Specific algorithms}
\label{sec:pr}
We now consider different approaches to solving
$\proc{Fit-Pois-Reg}({\bf A}, {\bm y})$, which, when inserted into
Algorithm~\ref{alg:bcd}, produce different Poisson NMF
algorithms. These algorithms are closely connected to existing
algorithms for Poisson NMF and/or the multinomial topic model (Table
\ref{table:algorithms}).

\subsubsection{Expectation maximization}

There is a long history of solving the Poisson regression problem
\eqref{eq:fit-pois-reg} by EM \citep{em, depierro-1993, em-book,
  krishnan-1995, lange-carson-1984, lucy-1974, meng-1997, molina-2001,
  richardson-1972, shepp-vardi-1982, vardi-1985, vardi-1993}. The EM
updates for this problem consist of iterating the following updates:
\begin{align}
\bar{z}_{ik} &= y_i a_{ik} b_k / \mu_i \label{eq:apr-E-step} \\
b_k &= \frac{\sum_{i=1}^n \bar{z}_{ik}}{\sum_{i=1}^n a_{ik}},
\label{eq:apr-M-step}
\end{align}
where $\bar{z}_{ik}$ represents a posterior expectation in an
equivalent augmented model (see Appendix \ref{sec:derivations}).

This EM algorithm is closely connected to the multiplicative
update rules for Poisson NMF \citep{lee-seung-2001}: combining the E
step \eqref{eq:apr-E-step} and M step \eqref{eq:apr-M-step} with the
substitutions used in Algorithm \ref{alg:bcd} yields
\begin{align}
h_{ik}^{\mathrm{new}} &\leftarrow h_{ik} \times
  \frac{\sum_{j=1}^m x_{ij} w_{jk} / \lambda_{ij}}
       {\sum_{j=1}^m w_{jk}} 
\label{eq:multiplicative-L} \\
w_{jk}^{\mathrm{new}} &\leftarrow w_{jk} \times
  \frac{\sum_{i=1}^n x_{ij} h_{ik} / \lambda_{ij}}
       {\sum_{i=1}^n h_{ik}},
\label{eq:multiplicative-F}
\end{align}
which are precisely the multiplicative updates for Poisson NMF. See
Appendix \ref{sec:derivations} for the derivation. Additionally,
applying $\proc{PNMF-to-MTM}$ to the
multiplicative updates
(\ref{eq:multiplicative-L}, \ref{eq:multiplicative-F}) recovers the EM
updates for the multinomial topic model \citep{asuncion-2009,
buntine-2002, gaussier-goutte-2005, hofmann-2001}. See Appendix
\ref{sec:derivations} for the derivation. Therefore, when 
$\proc{Fit-Pois-Reg}({\bf A}, {\bm y})$ is solved using EM,
Algorithm~\ref{alg:bcd} can be viewed as implementing ``stepwise''
variants of the multiplicative updates for Poisson NMF or EM for the
multinomial topic model (Table \ref{table:algorithms}).  By
``stepwise'', we mean that the update order suggested by
Algorithm \ref{alg:bcd} is to iterate the E and M steps for the first
row of ${\bf H}$, then for the second row of ${\bf H}$, and so on,
followed by updates to rows of ${\bf W}$. This is in contrast to a
typical EM algorithm in which all E-step updates are performed first,
then all M-step updates are performed.
\begin{table}[t]
\caption{Relationship between maximum-likelihood estimation algorithms for the
  additive Poisson regression model, Poisson NMF, and the multinomial
  topic model. Abbreviations used: EM = expectation
  maximization \citep{shepp-vardi-1982, hofmann-1999}, MU =
  multiplicative updates \citep{lee-seung-1999, lee-seung-2001}, CD =
  co-ordinate descent \citep{bouman-1996}, CCD = cyclic co-ordinate
  descent \citep{ccd}, SCD = sequential co-ordinate descent  \citep{scd}.
\label{table:algorithms}}
\begin{tabular}{@{}c@{\;\;\;}c@{\;\;\;}c@{}}
\toprule
additive Poisson regression & Poisson NMF & topic model \\
\midrule
EM & 
MU & 
EM \\[0.5ex]
CD & CCD, SCD & {\em none} \\
\bottomrule
\end{tabular}
\end{table}

\subsubsection{Co-ordinate descent}

Co-ordinate descent is an alternative to EM that iteratively optimizes
a single co-ordinate while the remaining co-ordinates are fixed
(see also \citealt{bouman-1996}). For the Poisson regression problem,
each 1-d optimization is straightforward to implement via Newton's
method,
\begin{equation}
b_k^{\mathrm{new}} \leftarrow \max\{0, b_k - \alpha_k g_k/q_k \},
\label{eq:cd}
\end{equation}
where $\alpha_k \geq 0$ is a step size that can be determined by a
line search or some other method, and $g_k$ and $q_k$ are the partial
derivatives with respect to the negative log-likelihood
$\ell_{\mathrm{PR}}({\bm b}) \colonequals -\log p_{\mathrm{PR}}({\bm
  y} \mid {\bf A}, {\bm b})$,
\begin{align}
g_k &\colonequals \frac{\partial \ell_{\mathrm{PR}}}{\partial b_k} 
= \sum_{i=1}^n a_{ik} \left(1 - \frac{y_i}{\mu_i}\right) \\
q_k &\colonequals \frac{\partial^2 \ell_{\mathrm{PR}}}{\partial b_k^2} 
= \sum_{i=1}^n \frac{y_i a_{ik}^2}{\mu_i^2}.
\end{align}

Several Poisson NMF algorithms, including cyclic co-ordinate descent (CCD)
\citep{ccd}, sequential co-ordinate descent (SCD) \citep{scd} and
scalar Newton (SN) \citep{hien-2020}, can be viewed as implementing
variants of this CD approach. That is, these approaches are essentially
Algorithm \ref{alg:bcd} in which $\proc{Fit-Pois-Reg}({\bf A}, {\bm
y})$ is solved by CD. The CCD and SCD methods appear to be independent
developments of the same or very similar algorithm; they both take a
full (feasible) Newton step, setting $\alpha_k = 1$ when $b_k
- \alpha_k g_k/q_k > 0$. By foregoing a line search to determine
$\alpha_k$, the update is not guaranteed to decrease the objective
$\ell_{\mathrm{PR}}({\bm b})$.
%
% \citep{nw}.
% 
% when ${\bm b}$ is far from a solution  
%
The SN method was developed to remedy this issue, with a step size scheme
that always produces a decrease while avoiding the expense of a line
search. However, \cite{hien-2020} compared SN with CCD
%
% (they did not compare SCD)
%
and found that CCD usually performed best in real data sets despite
not having a line search.

Although the CD approach is straightforward for Poisson NMF, it is not
straightforward for the multinomial topic model due to the sum-to-one
constraints.
%
% This probably explains why the CD approach has not been implemented
% for the multinomial topic model (Table \ref{table:algorithms}).
%

\section{Numerical experiments}
\label{sec:experiments}

To summarize, we have described two variants of Algorithm
\ref{alg:bcd} for Poisson NMF (Table~\ref{table:algorithms}): the
first fits an ``additive Poisson regression'' model using EM, and is
essentially the same as existing EM algorithms for Poisson NMF and the
multinomial topic model (including the Poisson NMF multiplicative
updates); the second uses co-ordinate descent (CD) to fit the additive
Poisson regression model, and has no equivalent among existing
algorithms for the multinomial topic model. In the remainder, we refer
to these two variants of Algorithm \ref{alg:bcd} as ``EM'' and ``CD''.

\begin{figure*}[t]
\centering{
\includegraphics[width=0.85\textwidth]{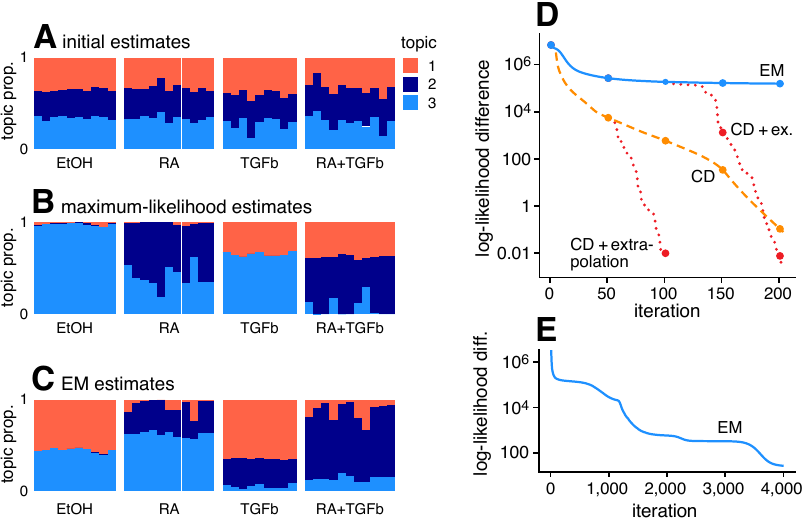}}
\caption{Results of fitting multinomial topic models to the MCF-7 data
  set \citep{mcf7} with $K = 3$. Plots A--C show estimates of the $41
  \times 3$ matrix ${\bf L}$: the initial estimates (obtained by
  running 4 EM updates); the MLE (obtained by running many CD updates,
  starting from the initial estimates); and the estimates obtained by
  running 200 EM updates starting from the initial estimates. Each
  estimate of ${\bf L}$ is visualized using a ``Structure plot''
  \citep{rosenberg-2002}, which is a stacked bar chart in which the
  bar heights are given by the elements of ${\bf L}$. Plots D, E show
  the improvement in the multinomial topic model fits over
  time. Multinomial topic model og-likelihoods are shown relative to
  the log-likelihood of the multinomial topic model at the MLE (B);
  points highest on the y-axis indicate the worst
  log-likelihoods. \label{fig:mcf7}}
\end{figure*}

We begin with an in-depth example on a real data set to illustrate the
differences between the EM and CD algorithms. The data for this
example are RNA-sequencing read counts for $n = 41$ samples and $m =
16,773$ genes from \cite{mcf7}. This data set provides a ground truth
of sorts for fitting the topic model: the data are gene expression
measurements taken after human MCF-7 cells were exposed to either
ethanol (EtOH), retinoic acid (RA), TGF-$\beta$, or the combination of
RA and TGF-$\beta$. Therefore, the topic model with $K = 3$ topics
should reflect the three exposures---EtOH, RA and TGF-$\beta$---and
samples in the combined exposure should be modeled as a combination of
the RA and TGF-$\beta$ topics. Indeed, the MLE we obtained by running
the Poisson NMF algorithm for a long time (with CD updates) largely
produced the expected result: the samples in the ethanol condition
were mostly represented by a single topic (the ``ethanol topic''); the
samples in the combination treatment were an even combination
of the RA and TGF-$\beta$ topics; and the samples exposed to either RA
and TGF-$\beta$ were represented as combinations of the ethanol and RA
topics or the ethanol and TGF-$\beta$ topics
(Fig.~\ref{fig:mcf7}B). The steps taken to prepare these data for
topic modeling are detailed in Appendix
\ref{sec:experiments-details}. The code implementing this experiment
is provided in a Zenodo repository \citep{zenodo}, and is available
online at
\url{https://github.com/stephenslab/fastTopics-experiments/}.

% This small simulation mirrors the experiments with real data
% (Sec.~\ref{sec:experiments}), but by avoiding the inherent
% complexities of real data the simulation allows us to better
% appreciate the differences in the algorithms' behaviour.

To compare the performance of EM and CD on this data set, we first
initialized the Poisson NMF parameters at random, then we ran 4 EM
updates to slightly improve upon this random initialization.  The
resulting initial estimate of ${\bf L}$ is shown in
Fig.~\ref{fig:mcf7}A.
%
% In our
% comparisons, we perform a small number of updates to obtain a good
% initialization. This reduces the possibility that different runs might
% find different local optima, which complicates the comparisons.
%
Next, starting from this initial estimate, we ran 200 EM updates or
200 CD updates. (``Update'' here means one iteration of the outer loop
of Algorithm \ref{alg:bcd}.) The CD updates produced estimates very
close to the MLE; the distance to the MLE in log-likelihood units was
just 0.079 (Fig. \ref{fig:mcf7}D). By contrast, the EM estimates
remained very far away from the MLE after 200 iterations, at a
distance of over 150,000 log-likelihood units
(Fig.~\ref{fig:mcf7}D). The EM estimates after 200 iterations
(Fig.~\ref{fig:mcf7}C) were also {\em qualitatively} very different
from the MLE. The EM estimates are arguably less interpretable
because they do not correspond as well to the exposures.

To rule out the possibility the EM was unlucky and had settled into a
different local maximum of the likelihood, we ran many more EM
updates. Eventually, EM recovered the same MLE
(Fig.~\ref{fig:mcf7}E). Therefore, the very slow progress of EM could
not be explained by having converged to a less optimal stationary
point. One could conclude from this comparison that good estimates
could be obtained simply by running the EM updates for a long
time. However, this is often impractical for larger data sets.
%
% Further, the fact that the EM updates eventually got ``unstuck''
% after running several thousand updates here does not guarantee that
% the same will occur in a different data set.
% 

Another algorithmic innovation we present here is the use of the
extrapolation method \citep{ang-2019} to accelerate convergence of the
Poisson NMF algorithm. The idea behind the extrapolation method, which
builds on the method of parallel tangents \citep{luenberger-ye}, is to
avoid the ``zigzagging'' behaviour of the block-coordinate updates by
iteratively adapting the step size according to the performance of the
extrapolated updates compared to the non-extrapolated updates. The
additional operations needed to implement the extrapolated updates
impose minimal overhead. The extrapolation method was originally
applied to Frobenius-norm NMF, and to our knowledge it has not been
used to accelerate algorithms for Poisson NMF, or for fitting topic
models. (More details on the extrapolation method and its
implementation for Poisson NMF are given in Appendix
\ref{sec:implementation}.)  To illustrate the benefits of
extrapolation, we activated the extrapolation method at iteration 50
of the CD algorithm. Doing so allowed the CD updates to recover the
MLE much more quickly than the non-extrapolated CD updates
(Fig.~\ref{fig:mcf7}D). Further, when we applied the extrapolated CD
updates to the EM estimates, they quickly ``rescued'' the poor EM
estimates (Fig.~\ref{fig:mcf7}D), further suggesting that the slow
progress of EM was not due to some fundamental difficulty of the
objective, but rather due to properties of the EM updates.

In summary, the results from this example suggest the potential for
NMF methods---in particular, Algorithm \ref{alg:bcd} with CD updates
plus extrapolation---to improve maximum-likelihood estimation for the
multinomial topic model. To assess this more systematically, we
performed comparisons of the EM and CD variants in a variety of data
sets (Table~\ref{table:datasets}): two text data sets
\citep{globerson-2007, newsgroups} that have been used to evaluate
topic modeling methods (e.g., \citealt{asuncion-2009, wallach-2006}); and
two single-cell RNA sequencing (scRNA-seq) data sets
\citep{montoro-2018, zheng-2017}. 
Appendix \ref{sec:experiments-details} gives additional details on
these data sets and the experiment setup.  All the algorithms
compared in these experiments were implemented in the fastTopics R
package.
%
% (see the supplementary ZIP file). 
%
These R implementations include the enhancements described in Appendix
\ref{sec:implementation} intended to make the algorithms more
efficient and numerically stable.  The code implementing these
experiments is provided in a Zenodo repository \citep{zenodo}, and is
available online at
\url{https://github.com/stephenslab/fastTopics-experiments/}.

\begin{table}[t]
\caption{Data sets used in the experiments. 
\label{table:datasets}}
\begin{tabular}{@{}l@{\;\;\;}c@{\;\;\;}c@{\;\;\;}c@{}}
\toprule
name & rows & columns & nonzeros \\
\midrule
NeurIPS           & 2,483  & 14,036 & 3.7\% \\
newsgroups        & 18,774 & 55,911 & 0.2\% \\
epithelial airway & 7,193  & 18,388 & 9.3\% \\
68k PBMC          & 68,579 & 20,387 & 2.7\% \\
\bottomrule
\end{tabular}
\end{table}

To reduce the possibility that multiple optimizations converge to
different local maxima of the likelihood, which could complicate the
comparisons, we first ran 1,000 EM updates---that is, 1,000 iterations
of the outer loop of Algorithm \ref{alg:bcd}---then we examined the
performance of the algorithms {\em after} this initialization phase.
Therefore, in our comparisons we assessed the extent to which the
different algorithms improved upon this initialization.
% 
% In addition to these quantitative
% performance comparisons, we also assessed whether different estimates
% qualitatively impact interpretation of the topics. 
%
Another issue was that it was not always practical to run the
optimization algorithms long enough to obtain an accurate
MLE. Therefore, instead of comparing the estimates to the MLE, like we
did in the example above, we used as a reference point the best
estimate (in log-likelihood) that was obtained.

\begin{figure*}[!t]
\centering{
\includegraphics[width=0.935\textwidth]{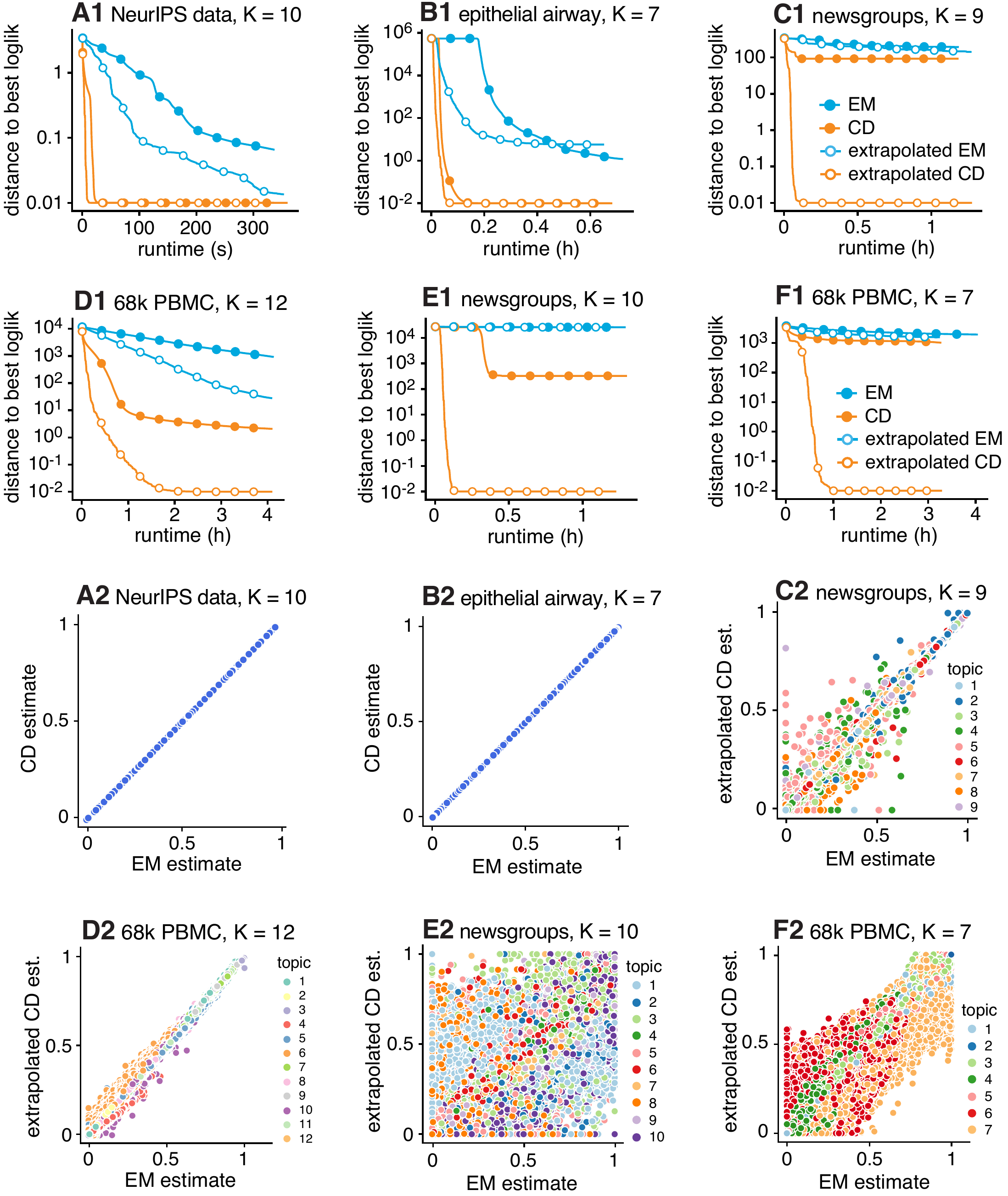}}
\caption{Selected results on fitting topic models using Poisson NMF
  algorithms. In A1--F1, multinomial topic model log-likelihoods are
  given relative to the best log-likelihood obtained among the four
  algorithms compared (EM and CD, with and without
  extrapolation). Log-likelihood differences less than 0.01 are shown
  as 0.01, and circles are drawn at intervals of 100 iterations. The
  1,000 EM iterations performed during the initialization phase are
  not shown. Plots A2--F2 compare the final estimates of ${\bf L}$
  from each of A1--F1. See also Figures
  \ref{fig:loglik-nips}--\ref{fig:kkt-pbmc68k} in the Appendix for
  additional results obtained with different settings of
  $K$.  \label{fig:results-main}}
\end{figure*}

\begin{figure*}[t]
\centering{
\includegraphics[width=\textwidth]{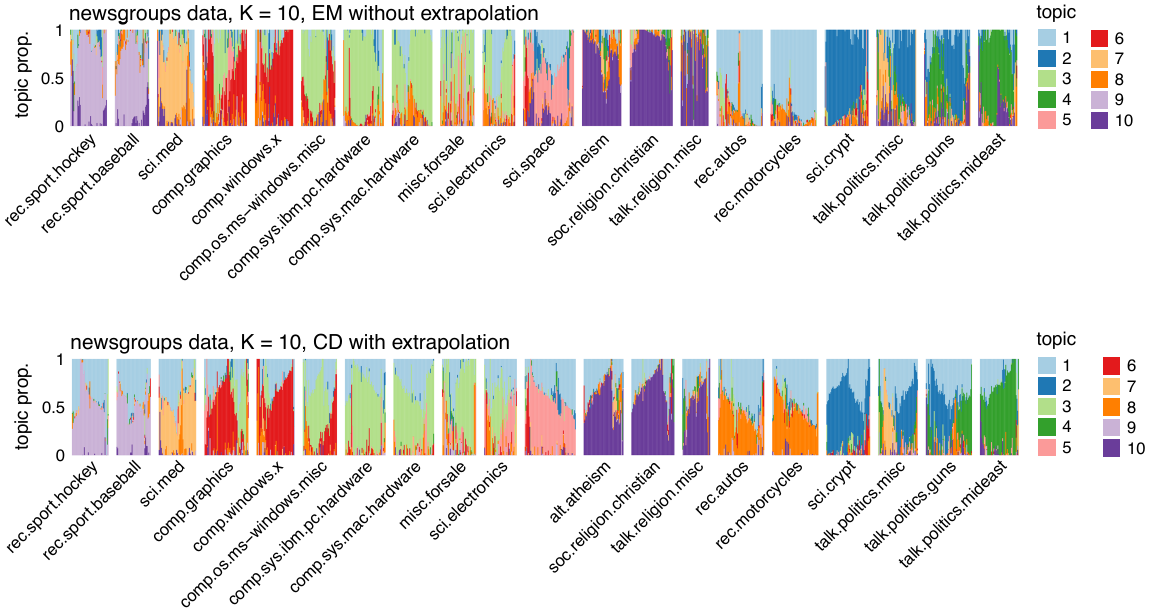}}
\caption{Estimates of ${\bf L}$ from the newsgroups data with $K = 10$
  obtained by running the EM updates without extrapolation (top) and
  the CD updates with extrapolation (bottom). The estimates of ${\bf
    L}$ are visualized using Structure plots. The documents are
  arranged by newsgroup to show the correspondence between the
  newsgroups and the topics. Note that the ordering of the documents
  within each newsgroup is not exactly the same in the top and bottom
  plots. See E1 and E2 in Fig. \ref{fig:results-main} for related
  results.}
\label{fig:structure-plots-newsgroups}
\end{figure*}

\begin{figure*}[t]
\centering{
\includegraphics[width=\textwidth]{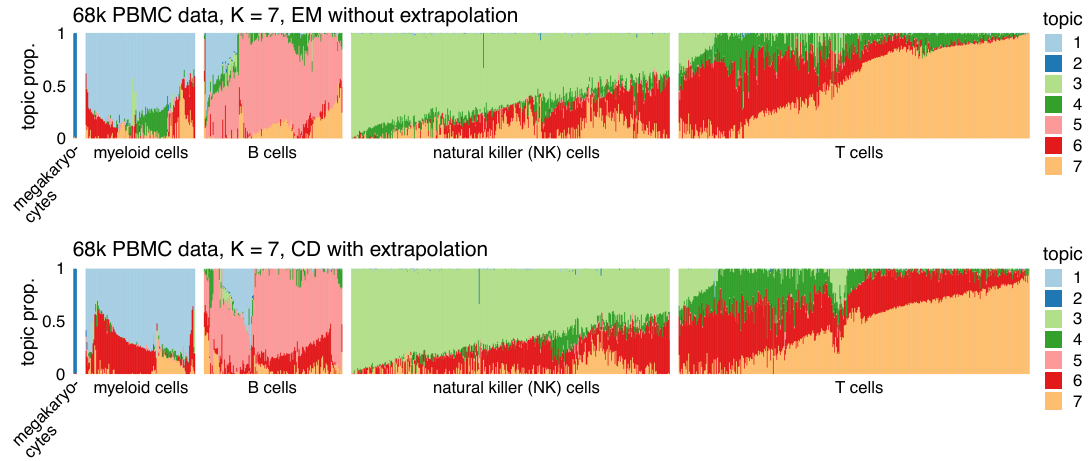}}
\caption{Estimates of ${\bf L}$ from the 68k PBMC data with $K = 7$
  obtained by running the EM updates without extrapolation (top) and
  the CD updates with extrapolation (bottom). The estimates of ${\bf
    L}$ are visualized using Structure plots.
  %
  % each bar in the Structure plot represents a cells,
  % and the bar heights of different colors give the topic
  % proportions. 
  %
  To facilitate comparison, the cells were split into 5 groups based
  on the CD estimates of ${\bf L}$; these groups roughly correspond to
  cell types (B cells, T cells, {\em etc}).  The ``T cells'' group was
  downsampled to better visualize the other groups. Note that the
  ordering of the cells within each grouping is not exactly the same
  in the top and bottom plots. See F1 and F2 in
  Fig. \ref{fig:results-main} for related results.}
\label{fig:structure-plots-pbmc68k}
\end{figure*}

Selected results of these comparisons are shown in
Fig. \ref{fig:results-main}, and more comprehensive results on all
four data sets, with $K$ ranging from 2 to 12, are given in the
Appendix (Figures \ref{fig:loglik-nips}--\ref{fig:kkt-pbmc68k}).  In
almost all cases, the extrapolated CD updates converged to an MLE at
least as fast as the other algorithms, and often much faster, or
produced the best fit within the allotted time.  The extrapolation
method generally helped convergence of CD, and sometimes helped
EM. Also, the per-iteration running time per was very similar in all
the algorithms. Beyond this, there was considerable variation in the
algorithms' performance among the different data sets and within each
data set at different settings of $K$. To make sense of the diverse
results, we distinguish three main patterns.

A1 and B1 in Fig. \ref{fig:results-main} illustrate the first pattern:
EM quickly progressed to a good solution, and so any improvements over
EM were small regardless of the algorithm used. Indeed, despite the
small improvements in log-likelihood obtained by the CD estimates in A1
and B1, the final EM and CD estimates were nearly indistinguishable
from each other (Fig. \ref{fig:results-main}, A2 and B2).

C1 and D1 in Fig. \ref{fig:results-main} illustrate the second
pattern: the initial 1,000 EM iterations were insufficient to
recover estimates close to an MLE, and running additional updates
sometimes substantially improved the fit. Among the four algorithms
compared, the extrapolated CD updates again provided the greatest
improvement in log-likelihood within the allotted time. And yet,
despite the considerable improvements in log-likelihood, the final
estimates did not change much (Fig. \ref{fig:results-main}, C2 and
D2). So while the CD updates can sometimes produce large gains in
computational performance, these gains do not always have a meaningful
impact on the topic modeling results.

E1 and F1 in Fig. \ref{fig:results-main} are examples of the third
pattern: the extrapolated CD updates not only produced estimates with
greatly improved log-likelihood, they also produced estimates that
were {\em qualitatively very different} (Fig. \ref{fig:results-main},
E2 and F2). In both this and the previous pattern, the EM updates
progressed slowly toward a solution. But whereas this slow progress
was benign in the previous examples, with little impact on the final
result,
%
% (and perhaps could even be beneficial by implicitly
% regularizing the estimates),
%
in these examples the slow progress of EM was in an area of the
likelihood that was very far away from an MLE. We also observed an
example of this pattern earlier in the MCF-7 data set
(Fig.~\ref{fig:mcf7}). In brief, the slow convergence of the EM
updates is sometimes benign, and sometimes not, but it is impossible
to know in advance which it is without making these
comparisons. Therefore, one way to avoid this problem is to use the CD
updates, which are generally better at not getting stuck in areas of
the likelihood far away from an MLE.

We also examined the topic model estimates in E and F to understand
how the improved estimates can affect our understanding of the
data. In the newsgroups data , topics 1 and 8 changed most between the
EM and CD estimates (Fig.~\ref{fig:structure-plots-newsgroups}): in
the EM estimates, the rec.auto and rec.motorcycle newsgroup
discussions were largely captured by topic 1, a topic that was also
shared by most other newsgroups; in the CD estimates, topic 8
distinguished rec.auto and rec.motorcycle from the other newsgroups,
and topic 1 was present more evenly in all the newsgroups.

In the 68k PBMC data (Fig. \ref{fig:structure-plots-pbmc68k}), there
were many differences between the EM and CD estimates, but the changes
to topic 4 most affect our understanding of these data: in the EM
estimates, the T cells shared topic 4 with a subset of myeloid
cells---suggesting some sort of pathway or gene expression program
common to myeloid and T cells---but in the improved CD estimates, this
connection between myeloid and T cells mostly disappeared, and topic 4
was mostly distinct to T cells.

% In topic 6 of the 68k
% PBMC data, the genes with the largest expression increases are
% ribosomal protein genes (Fig.~\ref{fig:genes-scatterplot-pbmc68k});
% abundance of ribosomal protein genes is known to be a major
% contributor to variation in PBMC scRNA-seq data sets
% \citep{freytag-2018}. Topics 3, 4, 6 and 7 capture heterogeneity in
% the largest cluster---the right-most cluster in
% Fig.~\ref{fig:structure-plots}F, mainly consisting of natural killer
% cells and T cells)---which brings to mind data set B in
% Sec.~\ref{sec:smallsim}. By contrast, topics 1, 2 and 5, which
% identify myeloid cells, CD34+ cells and B cells, respectively, are
% mostly independent, and the EM and CD estimates for these topics do
% not differ much (Fig.~\ref{fig:results-main}, Panel F2). These results
% suggest that interdependent topics can impede EM's convergence, and in
% these settings CD can offer large improvements.

\section{Concluding remarks}
\label{sec:discussion}

In this paper, we suggested a simple strategy for fitting topic models
by exploiting the equivalence of Poisson NMF and the multinomial topic
model: first fit a Poisson NMF, then recover the corresponding topic
model. To our knowledge, this equivalence, despite being informally
recognized early on in the development of these methods, has not been
previously exploited to fit topic models. The greatest improvements in
optimization performance were achieved when the Poisson NMF was
optimized using a simple co-ordinate descent (CD) algorithm. While the
CD algorithm may be simple, consider that, due to the ``sum-to-one''
constraints, it is not obvious how to implement CD for the multinomial
topic model.

For many statistical applications, point estimation such as
maximum-likelihood estimation will suffice when the main aim is to
learn a low-rank representation of the count data. (See
also \citealt{taddy-2012}.)
%
% for additional arguments in support of point estimation in topic
% models
%
Further, focussing on point estimation simplifies numerical
computation, allowing for simpler and more efficient algorithms that can be
quickly applied to large data sets.  We focussed on maximum-likelihood
estimation, but the ideas and algorithms presented here also apply to
MAP estimation with Dirichlet priors on ${\bf F}$. Extending these
ideas to improve variational inference algorithms for topic models
(e.g., LDA) may also be of interest. Given the success of the CD
approach, it may be fruitful to develop CD-based variational inference
algorithms for LDA and Poisson NMF \citep{gopalan-2015}.
%
% Simply initializing the VEM from the solution found by our approach
% may already improve fits.
%
That said, in many applications topic models are mainly used for
dimension reduction---the goal being to learn compact representations
of complex patterns---and in these applications maximum-likelihood or
MAP estimation may suffice.
%
% The speed of these algorithms on large data sets can be further
% improved by the use of ``online'' versions that exploit stochastic
% approximation ideas \citep{hoffman-2010, sato-2001}.  
%
Some topic modeling applications involve massive data sets that
require an ``online'' approach \citep{hoffman-2010, sato-2001,
hoffman2013stochastic, broderick2013streaming}.  Developing online
versions of our algorithms is straightforward in principle, although
online learning brings additional practical challenges, such as the
choice of learning rate.

% Of course there have been many important advances in topic
% models (e.g., correlated topic models, dynamic topic models,
% structural topic models). However the original (pLSI/LDA) topic
% model remains by far the most widely used topic model. Here we are
% are particularly interested in fitting topic models to single-cell
% data sets to identify structure in single-cell sequencing data.

% The intuition is that topic modeling algorithms may have
% difficulty breaking through interdependencies among topics when they
% occur, whereas the Poisson NMF optimization algorithms seem to cope
% better with these interdependencies. Since the use of priors
% typically does not introduce further interdependencies, a good ML
% estimate puts the algorithm in a good position to find good
% posterior estimates.

% Even in circumstances where maximum-likelihood estimation will
% not work well (``overfit''), an MLE can still
% be a useful starting point for MAP estimation or approximate
% posterior inference using variational methods. Discuss when
% maximum-likelihood estimation will not work well, and when it might
% fail ({\em i.e.}, ``overfit'').

It is well known that EM can suffer from slow convergence; recent
theoretical developments shed light on this slow convergence and the
conditions under which it occurs \citep{kunstner2021em, raz2020em}.
And so it is perhaps not surprising that the EM variant of the
Alternating Poisson Regression algorithm (Algorithm \ref{alg:bcd}) was
also very slow in some data sets.
%
% (The EM is the same as the multiplicative updates.)
%
However, we distinguished between two types of slow convergence:
benign slow convergence that occurs when the EM estimates are near an
MLE; and slow convergence far away from an MLE, which can result in EM
estimates that are very different from an MLE, and can affect how the
topics are interpreted. Several methods have been developed
specifically to accelerate EM \citep{zhou-lange-2011, varadhan-2008,
  daarem}, and therefore it would have been natural to apply these
methods here to improve the performance of the EM updates. We actually
tried two more recent acceleration methods---DAAREM \citep{daarem} and
the quasi-Newton method of \cite{zhou-lange-2011}---but in our tests
(results not shown) we found that both methods provided little
improvement over the unaccelerated EM. \cite{taddy-2012} also used a
quasi-Newton method to accelerate EM, but did not provide any results
to show that this was beneficial. The only acceleration method that
consistently improved performance was the extrapolation method of
\cite{ang-2019}.

% ...although we improved topic model fits compared with EM-like
% algorithms... It may be therefore fruitful to develop online
% versions of our algorithms.

\begin{appendices}

%\linenumbers

% \renewcommand{\thefigure}{A.\arabic{figure}}
% \renewcommand{\thesection}{Appendix~\Alph{section}:}
% \renewcommand{\thesubsection}{\Alph{section}.\arabic{subsection}}
\renewcommand{\thefigure}{\arabic{figure}}
\setcounter{figure}{4}

\section{Derivations and additional theory}
\label{sec:derivations}

\subsection{MAP estimation}
\label{sec:corollary-map}

\begin{corollary}[Relationship between MAP estimates for Poisson NMF and  
the multinomial topic model] \label{cor:map} \rm Let ${\hat{\bf H}}
  \in {\bf R}_{+}^{n \times K}, \hat{\bf W} \in {\bf R}_{+}^{m \times
    K}$ denote {\em maximum a posteriori} (MAP) estimates for the
  Poisson NMF model, in which elements of ${\bf W}$ are assigned
  independent gamma priors, $w_{jk} \sim \mathrm{Gamma}(a_{jk}, b_k)$,
  with $a_{jk} > 1$, $b_k > 0$, for $j = 1, \ldots, m$, $k = 1,
  \ldots, K$, and elements of ${\bf H}$ are assigned an (improper)
  uniform prior,
\begin{align} 
\label{eq:map_pnmf}
\hat{\bf H}, \hat{\bf W} \in \underset{{\bf H},\,{\bf W}}{\mathrm{argmax}} \;
 \pnmf(\X \mid {\bf H}, {\bf W}) \, p_{\mathrm{gam}}({\bf W}),
\end{align}
in which
\begin{equation*}
p_{\mathrm{gam}}({\bf W}) \colonequals \prod_{j=1}^m \prod_{k=1}^K 
\mathrm{Gamma}(w_{jk}; a_{jk}, b_k),  
\end{equation*}
and where $\mathrm{Gamma}(\theta; \alpha, \beta) \propto
\theta^{\alpha - 1} e^{-\beta\theta}$ denotes the probability density
of the gamma distribution with shape $\alpha$ and rate (inverse
scale) $\beta$. If $\hat{\L}, \hat{\F}$ are obtained by applying
$\proc{PNMF-to-MTM}$ to $\hat{\bf H}, \hat{\bf W}$, these are MAP
estimates for the multinomial topic model with independent Dirichlet
priors on the columns of ${\bf F}$, $f_{1k}, \ldots, f_{mk} \sim
\mathrm{Dirichlet}(a_{1k}, \ldots, a_{mk})$ and a uniform prior on
$\L$,
\begin{equation} 
\label{eq:map_topic}
\hat{\L},\hat{\F} \in \underset{\L,\,\F}{\mathrm{argmax}} \,
\ptop(\X \mid \L, \F) \, p_{\mathrm{dir}}({\bf F}),
\end{equation}
in which
\begin{equation*}
p_{\mathrm{dir}}({\bf F}) \colonequals \prod_{k=1}^K
\mathrm{Dirichlet}(f_{1k}, \ldots, f_{mk}; a_{1k}, \ldots, a_{mk}),
\end{equation*}
and $\mathrm{Dirichlet}(\theta_1, \ldots, \theta_d; \alpha_1,
\ldots, \alpha_d) \propto \theta_1^{\alpha_1 - 1} \cdots
\theta_d^{\alpha_d - 1}$ denotes the probability density of the
Dirichlet distribution with parameters $\alpha_1, \ldots, \alpha_d$.
Conversely, let $\hat{\L} \in {\bf R}_{\mathrm{row}}^{n \times K},
\hat{\F} \in {\bf R}_{\mathrm{col}}^{m \times K}$ denote multinomial
topic model MAP estimates \eqref{eq:map_topic}, set $\hat{s}_i = t_i =
\sum_{j=1}^m x_{ij}$, for $i = 1, \dots, n$, and set $\hat{u}_k =
\sum_{j=1}^m (a_{jk}-1) / b_k$, for $k = 1, \dots, K$. Then if
$\hat{\bf H}, \hat{\bf W}$ are obtained by applying
$\proc{MTM-to-PNMF}$ to $\hat{\L}, \hat{\F}, \hat{\bm s}, \hat{\bm
  u}$, these will be Poisson NMF MAP estimates \eqref{eq:map_pnmf}.
\end{corollary}

\begin{proof}
To prove this result, first note that minimizing the penalized
objective for the multinomial topic model, $\phi^{\star}({\bf X}; {\bf
  L}, {\bf F})$ (see eq.~\ref{eq:loss-mtm-penalized}), corresponds to
computing MAP estimates \eqref{eq:map_topic}. This penalized objective
can be written in the form of the unpenalized objective, $\phi({\bf
  X}; {\bf L}, {\bf F})$,
% $\mathrm{argmax}_{{\bf L}, \, {\bf F}} \, \phi^{\star}({\bf X}; {\bf
% L}, {\bf F}) = \mathrm{argmax}_{{\bf L}, \, {\bf F}} \,
% \phi(\tilde{\bf X}; \tilde{\bf L}, {\bf F})$
where $\tilde{\bf X}$ and $\tilde{\bf L}$ are matrices augmented with
an additional $K$ ``pseudodocuments'',
\begin{equation}
\tilde{\bf X} \colonequals
\left[\begin{array}{c}
{\bf X} \\
{\bf A}^T - 1
\end{array}\right], \quad
\tilde{\bf L} \colonequals
\left[\begin{array}{c}
{\bf L} \\
{\bf I}_K
\end{array}\right],
\label{eq:pseudodocuments}
\end{equation}
and where ${\bf A}$ is a $m \times K$ matrix with elements $a_{jk}$ and
${\bf I}_K$ is the $K \times K$ identity matrix.
%
% $\tilde{\bf X}$ is an $\tilde{n} \times m$ matrix, in which
% $\tilde{n} = n + K$, $\tilde{x}_{ij} = x_{ij}$ for all $i \leq n$, and
% for all rows $i > n$, with $i = n + k$, $\tilde{x}_{ij} = a_{jk} - 1$;
% $\tilde{\bf L}$ is an $\tilde{n} \times K$ matrix, in which
% $\tilde{l}_{ik} = l_{ik}$ for all $i \leq n$, and $\tilde{l}_{ik} =
% 1$, $\tilde{l}_{ik'} = 0$ for all $i > n$ such that $i = n + k$, $k'
% \neq k$.
%
In other words, each of the $K$ pseudodocuments is attributed entirely
to a single topic, and the Dirichlet prior parameters are treated as
``pseudocounts''.

Similarly, the Poisson NMF penalized loss function $\ell^{\star}({\bf
  X}; {\bf H}, {\bf W})$---in which minimizing this loss function
corresponds to computing Poisson NMF MAP estimates
\eqref{eq:map_pnmf}---can be rewritten in the form of the unpenalized
Poisson NMF objective function, $\ell({\bf X}; {\bf H}, {\bf W})$. To
show this, we employ a change of variables that rescales the columns
of ${\bf H}$ and ${\bf W}$, ${\bf W}' \leftarrow {\bf W} {\bf U}^{-1}$,
${\bf H}' \leftarrow {\bf H} {\bf U}$, where ${\bf U} \colonequals
\mathrm{diag}({\bm u})$, and the columns of ${\bf W}'$ are constrained
to sum to one, $\sum_{j=1}^m w_{jk}' = 1$, $k = 1, \ldots, K$.
%
% replace $w_{jk}$ in the Poisson NMF model with $u_k w_{jk}'$, then
% replace $h_{ik} u_k$ with $h_{ik}'$.
% 
With this change of variables, the penalized loss function can be
rewritten as
\begin{align*}
\ell^{\star}({\bf X}; {\bf H}', {\bf W}') &=
% \phi({\bf H}', {\bf W}') + \|{\bf H}'({\bf W}')^T\|_{1,1}
\ell({\bf X}; {\bf H}', {\bf W}')
\nonumber \\
& \quad - \sum_{j=1}^m \sum_{k=1}^K (a_{jk} - 1) \log (u_k w_{jk}')
\nonumber \\
& \quad + \sum_{k=1}^K b_k u_k.
\end{align*}
With this objective, one can independently solve for each $u_k$, with
analytic solution $\hat{u}_k = \sum_{j=1}^m (a_{jk} - 1)/b_k$
that does not depend on the other parameters. Therefore, we focus on
the part of the loss function that depends on ${\bf H}', {\bf W}'$,
that is,
\begin{align*}
\ell^{\star}({\bf X}; {\bf H}', {\bf W}') &=
% \phi({\bf X}; {\bf H}', {\bf W}') + \|{\bf H}'({\bf W}')^T\|_{1,1}
\ell({\bf X}; {\bf H}', {\bf W}')
\nonumber \\
& \quad - \sum_{j=1}^m \sum_{k=1}^K (a_{jk} - 1) \log w_{jk}'
+ \mbox{const},
\end{align*}
where the ``const'' is a placeholder for terms that do not depend on
${\bf H}'$ or ${\bf W}'$. This can written as $\ell^{\star}({\bf X};
{\bf H}', {\bf W}') = \ell(\tilde{\bf X}; \tilde{\bf H}'; {\bf W}') +
\mbox{const}$, where $\tilde{\bf X}$ is the data matrix augmented with
pseudocounts \eqref{eq:pseudodocuments}, and $\tilde{\bf H}'
\colonequals \left[\begin{array}{c} {\bf H}' \\ {\bf
      I}_K \end{array}\right]$ is the matrix ${\bf H}'$ augmented with
pseudodocuments. Finally, we revert back to the original
parameterization, $\tilde{\bf H} \leftarrow \tilde{\bf H}' \hat{\bf
  U}^{-1}, \quad {\bf W} \leftarrow {\bf W}' \hat{\bf U}$, where
$\hat{\bf U} \colonequals \mathrm{diag}(\hat{\bm u})$.
%
% $\tilde{h}_{ik} \leftarrow \tilde{h}_{ik}'/\hat{u}_k$, $w_{jk}
% \leftarrow \hat{u}_k w_{jk}'$.
%
% The end result is loss function $\ell(\tilde{\bf X}; \tilde{\bf H},
% {\bf W})$, with the additional constraints that $\tilde{h}_{ik'} = 0$
% (equivalently, $\tilde{l}_{ik'} = 0$) whenever $i > n$, $i = n + k$
% and $k' \neq k$.
%

To summarize, this shows that MAP estimation (\ref{eq:map_pnmf},
\ref{eq:map_topic}) can be reduced to MLE estimation
(\ref{eq:mle-pnmf}, \ref{eq:mle-multinomial-topic-model}) when ${\bf
  X}$ is replaced with $\tilde{\bf X}$, and ${\bf H}$ is replaced with
$\tilde{\bf H}$; that is,
\begin{align}
& \mathrm{argmax}_{{\bf H},\,{\bf W}} \;
\pnmf(\X \mid {\bf H}, {\bf W}) \, p_{\mathrm{gam}}({\bf W}) \nonumber \\
&\quad = \mathrm{argmax}_{{\bf H}, \, {\bf W}} \;
\pnmf(\tilde{\X} \mid \tilde{\bf H}, {\bf W}) \\
& \mathrm{argmax}_{\L,\,\F} \;
\ptop(\X \mid \L, \F) \, p_{\mathrm{dir}}({\bf F}) \nonumber \\
&\quad = \mathrm{argmax}_{{\bf L}, \, {\bf F}} \;
\ptop(\tilde{\X} \mid \tilde{\L}, \F).
\end{align}
Therefore, we can apply Corollary \ref{cor:mle} to prove Corollary
\ref{cor:map}.
\end{proof}

\subsection{EM algorithms}
\label{sec:em}

\subsubsection{EM for the additive Poisson regression model}
\label{sec:em-apr}

Here we rederive the basic EM algorithm \citep{lange-carson-1984,
  shepp-vardi-1982, vardi-1985} for fitting the additive Poisson
  regression model \eqref{eq:apr}. To do so, we first introduce a
  data-augmented version of the Poisson regression model:
\begin{equation}
\begin{aligned}
z_{ik} &\sim \mathrm{Pois}(a_{ik} b_k) \\
y_i &= \textstyle \sum_{k=1}^K z_{ik}.
\end{aligned}
\end{equation}
Under this data-augmented model, the expected complete log-likelihood
is
\begin{align}
E[\log p({\bm y}, {\bm z} \mid {\bf A}, {\bm b})] &= 
\sum_{i=1}^n \sum_{k=1}^K \bar{z}_{ik} \log(a_{ik} b_k)
\nonumber \\
& \quad - \sum_{i=1}^n \sum_{k=1}^K a_{ik} b_k + \mbox{const},
\label{eq:apr-augmented}
\end{align}
where ``const'' includes additional terms in the likelihood that do
not depend on ${\bm b}$, and $\bar{z}_{ik} = E[z_{ik}]$ is the
expected value of $z_{ik}$ with respect to the posterior $p({\bm z}
\mid {\bf A}, {\bm b}, {\bm y})$.

Using this data-augmented model, the M step \eqref{eq:apr-M-step} is
derived by taking the partial derivative of \eqref{eq:apr-augmented}
with respect to $b_k$, and solving for $b_k$.
% \begin{equation}
% b_k = \frac{\sum_{i=1}^n \bar{z}_{ik}}{\sum_{i=1}^n a_{ik}}.
% \end{equation}
The E step \eqref{eq:apr-M-step} involves computing posterior
expectations at the current ${\bm b} = (b_1, \ldots, b_K)$. The
posterior distribution of $z_i = (z_{i1}, \ldots, z_{iK})$ is
multinomial with $y_i$ trials and multinomial probabilities $p_{ik}
\propto a_{ik} b_k$. Therefore, the posterior expected value of
$z_{ik}$ is
\begin{equation}
\bar{z}_{ik} = y_i p_{ik} = y_i a_{ik} b_k / \mu_i.
\label{eq:apr-E-step-expanded}
\end{equation}

The EM algorithm iterates the E and M steps until some stopping
criterion is met. Alternatively, the E and M steps can be combined,
yielding the update
\begin{equation}
b_k^{\rm new} \leftarrow
b_k \times \frac{\sum_{i=1}^n a_{ik} y_i/\mu_i}{\sum_{i=1}^n a_{ik}}.
\label{eq:apr-em-update}
\end{equation}

\subsubsection{Alternative EM Algorithm for additive Poisson regression}
\label{sec:em-apr-2}

The additive Poisson regression model \eqref{eq:apr} is equivalent to
a {\em multinomial mixture model} by a simple reparameterization.
%
% For practical implementation (with finite precision arithmetic), the
% EM updates for the multinomial mixture model are more convenient, so
% we use the following reparameterization to implement EM for the
% additive Poisson regression model.
%
The multinomial mixture model is
\begin{equation}
\begin{array}{r@{\;}c@{\;}l}
y_1, \ldots, y_n &\sim& \mathrm{Multin}(t, {\bm\pi}), \\
\pi_i &=& \sum_{k=1}^K a_{ik}' b_k',
\end{array}
\label{eq:multmix}
\end{equation}
in which ${\bf A}' \in {\bf R}_{+}^{n \times K}$, ${\bm y} = (y_1,
\ldots, y_n) \in {\bf R}_{+}^n$, ${\bm\pi} = (\pi_1, \ldots, \pi_n)$
and $t = \sum_{i=1}^n y_i$. To ensure that the $\pi_i$'s are
probabilities, we require $b_k' \geq 0$, $a_{ik}' \geq 0$,
$\sum_{k=1}^K b_k' = 1$, $\sum_{i=1}^n a_{ik}' = 1$. The multinomial
mixture model is a reparameterization of the Poisson regression model
\eqref{eq:apr} that preserves the likelihood; that is,
\begin{equation}
\prod_{i=1}^n \mathrm{Pois}(y_i; \mu_i) =
\mathrm{Multin}({\bm y}; t, {\bm\pi}) 
\, \mathrm{Pois}(t; s),
\end{equation}
in which the right-hand side quantities ${\bf A}', {\bm b}', s$ are
recovered from the left-hand side quantities ${\bf A}, {\bm b}$ as
follows:
\begin{equation}
\begin{aligned}
u_k     &\leftarrow \textstyle \sum_{i=1}^n a_{ik} \\
s       &\leftarrow \textstyle \sum_{k=1}^K b_k u_k \\
a_{ik}' &\leftarrow a_{ik} / u_k \\
b_k'    &\leftarrow b_k u_k / s.
\end{aligned}
\end{equation}

The EM algorithm for fitting the multinomial mixture model
consists of iterating the following E and M steps:
\begin{align}
p_{ik} &= \frac{a_{ik}' b_k'}{\sum_{j=1}^K a_{ij}' b_j'} \\
b_k' &= \frac{1}{t} \sum_{i=1}^n y_i p_{ik}.
\end{align}
Once the multinomial mixture model parameters $b_1', \ldots, b_K'$
have been updated by performing one or more EM updates, the Poisson
regression model parameters $b_1, \ldots, b_K$ are recovered as $b_k =
t b_k'/u_k$, using the MLE of $s$, $s = t$.

\subsubsection{Multiplicative updates for Poisson NMF}
\label{sec:em-poisson-nmf}

Having derived EM for the additive Poisson regression model in the
previous section, we can use this result to re-derive the
multiplicative updates for Poisson NMF \citep{lee-seung-2001} by
making substitutions ${\bf A} \leftarrow {\bf W}$, ${\bm y} \leftarrow
{\bm x}_i$, ${\bm b} \leftarrow {\bm h}_i$ in
\eqref{eq:apr-em-update}, where ${\bm h}_i$ denotes a row of ${\bf H}$
and ${\bm x}_i$ denotes a row of ${\bf X}$, the update becomes the
Poisson NMF multiplicative update for row $i$ of ${\bf H}$;
% \begin{equation}
% h_{ik}^{\rm new} \leftarrow h_{ik} \times
%   \frac{\sum_{j=1}^m x_{ij} w_{jk} / \lambda_{ij}}
%        {\sum_{j=1}^m w_{jk}}.
% \end{equation}
similarly, making substitutions ${\bf A} \leftarrow {\bf H}$, ${\bm y}
\leftarrow {\bm x}_j$, ${\bm b} \leftarrow {\bm w}_j$ in
\eqref{eq:apr-em-update}, where ${\bm w}_j$ denotes a row of ${\bf W}$
and ${\bm x}_j$ denotes a column of ${\bf X}$, the update becomes the
Poisson NMF multiplicative update for row $j$ of ${\bf W}$.
% \begin{equation}
% w_{jk}^{\rm new} \leftarrow w_{jk} \times
%   \frac{\sum_{i=1}^n x_{ij} h_{ik} / \lambda_{ij}}
%        {\sum_{i=1}^n h_{ik}}.
% \end{equation}
% These are precisely the ``multiplicative update rules'' of
% \cite{lee-seung-2001}.

\subsubsection{EM for the multinomial topic model}
\label{sec:em-plsi}

Here we derive EM for the multinomial topic model
\citep{hofmann-2001}, and connect EM for the
multinomial topic model and the multiplicative updates for Poisson
NMF. The EM algorithm for the multinomial topic model is based on a
data-augmented version of the topic model \citep{blei-2003},
\begin{equation}
\begin{aligned}
p(z_{it} = k \mid {\bf L}) &= l_{ik} \\
p(d_{it} = j \mid \F, z_{it} = k) &= f_{jk},
\end{aligned}
\label{eq:multinomial-topic-model-augmented}
\end{equation}
where $d_{it} \in \{1, \ldots, K\}$, and the data are $d_{it} \in \{1,
\ldots, m\}$, $t = 1, \ldots, n_i$, in which $n_i$ is the size of
document $i$. Summing over the topic assignments $z_{ij}$ recovers the
multinomial topic model \eqref{eq:multinomial-topic-model}, in which
the word counts are recovered as $x_{ij} = \sum_{t = 1}^{n_i}
\delta_j(d_{it})$.

% up to a constant of proportionality; that is,
% \begin{equation*}
% p({\X} \mid \L^{\star}, {\bf  F}^{\star}) 
% \propto {\textstyle \int p({\bm w}, {\bm z} \mid \L^{\star}, \F^{\star}) \,
%   d{\bm z}} \\
% = \prod_{i=1}^n \prod_{r=1}^{t_i} 
%    \sum_{z_{ir} = 1}^K p(w_{ir} \mid \F^{\star}, z_{ir}) \,
%    p(z_{ir} \mid {\bm l}_i^{\star}),
% \end{equation*}

The E step consists of computing the posterior expected values for the
latent topic assignments $z_{ij}$,
\begin{align}
p_{ijk} &\colonequals
p(z_{ij} = k \mid {\X}, \L, \F) =
l_{ik} f_{jk}/\pi_{ij}.
\label{eq:plsi-E-step}
\end{align}
The M step for the topic proportions $l_{ik}$ and word
frequencies $f_{jk}$ is
\begin{align}
l_{ik} &= \sum_{j=1}^m x_{ij} p_{ijk} / n_i
\label{eq:plsi-M-step-loadings} \\
f_{jk} &\propto \sum_{i=1}^n x_{ij} p_{ijk}.
\label{eq:plsi-M-step-factors}
\end{align}

Combining the E and M steps, we obtain the following updates:
\begin{align}
l_{ik}^{\rm new} &\leftarrow
\frac{l_{ik}}{t_i} \sum_{j=1}^m x_{ij} f_{jk}/\pi_{ij}
\label{eq:plsi-em-loadings} \\
f_{jk}^{\rm new} &\leftarrow \frac{f_{jk}}{\xi_k}
\sum_{i=1}^n x_{ij} l_{ik} / \pi_{ij}.
\label{eq:plsi-em-factors}
\end{align}
Here, $\xi_k > 0$ is a normalizing factor that ensures that
$\sum_{j=1}^m f_{jk}^{\rm new} = 1$. To connect to Poisson NMF, these
updates can also be derived by applying $\proc{PNMF-to-MTM}$ and
afterward its inverse to the Poisson NMF multiplicative updates
(\ref{eq:multiplicative-L}, \ref{eq:multiplicative-F}).

\subsection{KKT conditions}
\label{sec:kkt}

The first-order KKT conditions for the Poisson NMF optimization
problem \eqref{eq:poisson-nmf-problem} are
\begin{align}
\nabla_{\bf H} 
\mathcal{L}({\bf X}; {\bf H}, {\bf W}, {\bf \Gamma}, {\bf \Omega}) &=
{\bm 0} \label{eq:kkt-1} \\
\nabla_{\bf W} 
\mathcal{L}({\bf X}; {\bf H}, {\bf W}, {\bf \Gamma}, {\bf \Omega}) &=
{\bm 0} \label{eq:kkt-2} \\
{\bf \Gamma} \odot {\bf H} &= {\bm 0} \label{eq:kkt-3} \\
{\bf \Omega} \odot {\bf W} &= {\bm 0} \label{eq:kkt-4} 
\end{align}
in which $\mathcal{L}({\bf X}; {\bf H}, {\bf W}, {\bf \Gamma},
{\bf \Omega})$ denotes the Lagrangian function,
\begin{align*}
\mathcal{L}({\bf X}; {\bf H}, {\bf W}, {\bf \Gamma}, {\bf \Omega}) 
\colonequals\;& \ell({\bf X}; {\bf H}, {\bf W}) 
- \| {\bf \Gamma} \odot {\bf H} \|_{1,1}
\nonumber \\ & 
- \| {\bf \Omega} \odot {\bf W} \|_{1,1}.
\end{align*}
To define the Lagrangian function, we have introduced two matrices of
Lagrange multipliers, ${\bf \Gamma} \in {\bf R}_{+}^{n \times K}$ and
${\bf \Omega} \in {\bf R}_{+}^{m \times K}$, associated with the
non-negativity constraints ${\bf H} \geq {\bm 0}$ and ${\bf W} \geq
{\bm 0}$. Combining \eqref{eq:kkt-1} and \eqref{eq:kkt-2}, we obtain
\begin{equation}
\begin{aligned}
{\bf \Omega} &= (1 - {\bf U})^T{\bf H}  \\ 
{\bf \Gamma} &= (1 - {\bf U}){\bf W},
\end{aligned}
\label{eq:kkt-simplified}
\end{equation}
in which ${\bf U}$ is an $n \times m$ matrix with entries $u_{ij} =
x_{ij} / \lambda_{ij}$. See also \cite{dhillon-2005} and
\cite{fevotte-2011} for generalizations of these KKT conditions.

\section{Algorithm implementation and enhancements}
\label{sec:implementation} 

\subsection{Extrapolated updates}
\label{sec:extrapolation}

To accelerate convergence of the EM and CD updates, we used the
extrapolation method of \cite{ang-2019}.
%
%  which is based on the method of parallel
% tangents \citep{luenberger-ye}.
%
In brief, at iteration $t$, the extrapolated update is
\begin{equation}
\begin{aligned}
{\bf H}^{\mathrm{ext}} &\leftarrow P_{+}[{\bf H}^{\mathrm{new}} + 
\beta^{(t)}({\bf H}^{\mathrm{new}} - {\bf H}^{(t-1)})] \\
{\bf W}^{\mathrm{ext}} &\leftarrow P_{+}[{\bf W}^{\mathrm{new}} + 
\beta^{(t)}({\bf W}^{\mathrm{new}} - {\bf W}^{(t-1)})],
\end{aligned}
\end{equation}
where ${\bf H}^{\mathrm{new}}$ is a new estimate obtained by solving
$\proc{Fit-Pois-Reg}({\bf W}^{(t-1)}, {\bm x}_i)$ for each $i = 1,
\ldots, n$, ${\bf W}^{\mathrm{new}}$ is a new estimate obtained by
$\proc{Fit-Pois-Reg}({\bf H}^{\mathrm{ext}}, {\bm x}_j)$ for each $j =
1, \ldots, m$, $P_{+}({\bf A})$ is the projection of matrix ${\bf A}$
onto the non-negative orthant, $P_{+}({\bf A})_{ij} \colonequals
\max\{0,({\bf A})_{ij}\}$, and $\beta^{(t)} \in [0,1]$ is a parameter
that interpolates between the new estimate and the estimate from the
previous iteration. (Note that setting $\beta^{(t)} = 0$ recovers the
update with no extrapolation.) Although \cite{ang-2019} developed the
extrapolation method for Frobenius-norm NMF, our initial trials showed
that it also worked well for Poisson NMF. It also worked better than
the other acceleration schemes we tried, the damped Anderson (DAAREM)
method of \cite{daarem} and the quasi-Newton acceleration method of
\cite{zhou-lange-2011}.

%
% leading to improved convergence with little
% extra computational effort.
%

% Store the value of the objective (loss) function at the current
% iterate (Fn, Ly).
% loss0.fnly <- fit$loss.fnly
% fit$loss.fnly <- sum(cost(X,fit$Ly,t(Fn),control$eps))
% If the solution improves the "current best" estimate, update the
% current best estimate using the non-extrapolated estimates of the
% factors (Fn) and the extrapolated estimates of the loadings (Ly).
% if (fit$loss.fnly < fit$loss) {
%   fit$F    <- Fn
%   fit$L    <- fit$Ly 
% }

%
% To assess the benefits of extrapolation for fitting Poisson NMF and
% topic models, in our experiments we compared the performance of the
% Poisson NMF optimization algorithms with and without extrapolation.
%

\subsection{Other enhancements and implementation details}

Here we detail other steps taken to speed up computation and improve
numerical stability of the Poisson NMF optimization algorithms.

{\bf Computations with sparse data.} Topic modeling data sets often
have very high levels of sparsity; that is, most of the counts
$x_{ij}$ are zero. We therefore used sparse matrix computation
techniques to reduce computational effort for sparse data sets. To
illustrate the importance of sparse computations, consider computing
the Poisson NMF loss function $\ell({\bf X}; {\bf H}, {\bf W})$ when
$\X$ is sparse. Once the $\|{\bf H} {\bf W}^T\|_{1,1}$ term is
computed, which requires $O((n + m)K)$ operations, computing
$\phi({\bf X}; {\bf H}, {\bf W})$ requires and additional $O(NK)$
operations, where $N$ is the number of nonzeros in ${\bf X}$, because
terms in the sum corresponding to $x_{ij} = 0$ can be
ignored. Therefore, for sparse $\X$ the time complexity of computing
$\ell({\bf X}; {\bf H}, {\bf W})$ is $O((N + n + m)K)$. Without
considering sparsity, the time complexity is $O(nmK)$, which is much
greater than $O((N + n + m)K)$ when $n \times m \gg N$. Similar logic
applies to other computations such as gradients and EM
(multiplicative) updates.

{\bf Incomplete optimization of subproblems.} In practice, accurately
solving each subproblem $\proc{Fit-Pois-Reg}(\L, {\bm x}_j)$ and
$\proc{Fit-Pois-Reg}(\F, {\bm x}_i)$ may not be necessary,
particularly in initial stages when ${\bf H}$ and ${\bf W}$ change a
lot from one iteration to the next. Incompletely solving the
subproblems has been shown to work well for Frobenius-norm
NMF \citep{gillis-2012, ccd, kim-2014}. Therefore, instead of running
the EM or CD algorithm to convergence, we stopped the optimization
early when the number of iterations exceeded some limit. We found that
performing at most 4 EM or CD updates worked well in practice when 
initialized to the estimate from the previous (outer-loop)
iteration. We write this incomplete optimization as
$\proc{Fit-Pois-Reg}({\bf A}, {\bf y}, {\bf b}_0)$, where ${\bm b}_0$
makes explicit the dependence on an initial estimate. Therefore, when
the subproblems are solved incompletely, $\proc{Fit-Pois-Reg}({\bf
W}^{(t-1)}, {\bm x}_i)$ is replaced with $\proc{Fit-Pois-Reg}({\bf
W}^{(t-1)}, {\bm x}_i, {\bm h}_i^{(t-1)})$ in Algorithm~\ref{alg:bcd},
and $\proc{Fit-Pois-Reg}({\bf H}^{(t)}, {\bm x}_j)$ is replaced with
$\proc{Fit-Pois-Reg}({\bf H}^{(t)}, {\bm x}_j, {\bm w}_j^{(t-1)})$.

{\bf Parallel computations.} We used Intel Threading Building Blocks
(TBB) multithreading
\citep{tbb} to solve $\proc{Fit-Pois-Reg}({\bf W}, {\bm x}_i)$, $i =
1, \ldots, n$, and $\proc{Fit-Pois-Reg}({\bf H}, {\bm x}_j)$, $j = 1,
\ldots, m$, in parallel.

% {\bf Initialization.} To obtain a good initialization, we used the
% Topic-SCORE algorithm \citep{ke-2019} to estimate ${\bf W}$, then we
% ran 10 CD updates of ${\bf H}$. The Topic-SCORE algorithm was very
% fast, and typically ran in less time than it took to run a single
% (outer-loop) iteration of Algorithm \ref{alg:bcd}.

{\bf Refinement of CD updates.} Since the CD updates were performed
without a line search, we found that the CD updates sometimes failed
to improve the objective when the iterate was far away from a
solution. Therefore, to reduce the failure rate of the CD updates, we
performed a single EM update prior to running each CD update.

{\bf Improved convergence guarantees.} Following \cite{gillis-2012},
any parameter estimates that fell below $10^{-15}$ were set to this
value.

{\bf Rescaled updates.} To improve numerical stability of the updates,
as others have done (e.g., \citealt{lee-seung-1999}), we rescaled ${\bf
W}$ and ${\bf H}$ after each full update.  Specifically, we rescaled
the matrices so that the column means of ${\bf W}$ were equal to the
column means of ${\bf H}$. Note that the Poisson rates $\lambda_{ij} =
({\bf H}{\bf W}^T)_{ij}$ and the Poisson NMF loss function $\ell({\bf
X}; {\bf H}, {\bf W})$ are invariant to this rescaling.

{\bf Assessing convergence.} We used two measures to assess
convergence of the iterates: (1) the change in the loss function
$\ell({\bf X}; {\bf H}, {\bf W})$, which is the same as the change in
the Poisson NMF log-likelihood; and (2) the maximum residual of the
KKT conditions.
% 
% See Appendix \ref{sec:kkt} for a derivation of the KKT conditions.
%

\section{Details of the numerical experiments}
\label{sec:experiments-details}

% These sections give additional details on the experiments: the data
% sets and how they were prepared; details on the computing environment;
% (Sec.~\ref{sec:computing-environment}); and details on the software
% and code used.

\subsection{Data sets}
\label{sec:datasets}

% See Table \ref{table:datasets} for summary of data sets used in the
% numerical experiments.

The NeurIPS \citep{globerson-2007} and newsgroups \citep{newsgroups}
data sets are word counts extracted from, respectively, 1988--2003
NeurIPS (formerly NIPS) papers and posts to 20 different
newsgroups. 
%
% Both data sets have been used to evaluate topic modeling
% methods (e.g., \citealt{asuncion-2009, wallach-2006}).  
%
The data sets were retrieved
from \url{http://ai.stanford.edu/~gal/data.html} and
\url{http://qwone.com/~jason/20Newsgroups}.
Documents with fewer than 2 nonzero word counts were removed.
% 
% Both datasets were
% distributed without a license.\footnote{Private communications with
% Jason Rennie and Gal Chechik.}
% 

The MCF-7 data are RNA-sequencing read counts from human MCF-7
cells \citep{mcf7}. These data were downloaded from the Gene Expression
Omnibus (GEO) website, accession GSE152749.

The epithelial airway and 68k PBMC data sets are UMI (unique molecular
identifier) counts from single-cell RNA sequencing experiments in
trachea epithelial cells in C57BL/6 mice \citep{montoro-2018} and in
``unsorted'' human peripheral blood mononuclear cells (PBMCs) [Fresh
68k PBMC Donor A] \citep{zheng-2017}. The epithelial airway data were
downloaded from GEO, accession
GSE103354. Specifically, we downloaded file {\tt
GSE103354\_Trachea\_\-drop\-let\_\-UMI\-counts.txt.gz}.  
Genes that
were not expressed in any of the cells were removed. 
%
% These data are publicly available on the GEO website and as such
% follow the NCBI copyright and data usage policies. These data were
% obtained under IRB-approved protocols \citep{montoro-2018}.
%
% The 68k PBMC data have been used to benchmark methods for single-cell
% RNA-seq data (e.g., \citealt{abdelaal-2019, diaz-mejia-2019,
%   townes-2019, gom_de}). 
%
For the 68k PBMC data, we downloaded the ``Gene/cell matrix
(filtered)'' {\tt tar.gz} file for the Fresh 68k PBMCs (Donor A) data
set from the 10x Genomics website
(\url{https://www.10xgenomics.com/datasets}). Genes that were not
expressed in any of the cells were removed.
%
% These data are distributed under the CC BY 4.0 license and were
% obtained under IRB-approved protocols \citep{zheng-2017}.
%

All data sets except the MCF-7 data set were stored as sparse
$n \times m$ count matrices ${\bf X}$, where $n$ is the number of
documents or cells, and $m$ is the number of words or genes. The MCF-7
data were not sparse, so they were stored as a dense matrix. The data
processing scripts are provided in the Zenodo repository \citep{zenodo}.

\subsection{Computing environment}
\label{sec:computing-environment}

All computations on real data sets were run in R 3.5.1 \citep{R},
linked to the OpenBLAS 0.2.19 optimized numerical libraries, on Linux
machines (Scientific Linux 7.4) with Intel Xeon E5-2680v4
(``Broadwell'') processors. For running the Poisson NMF optimization
algorithms, which included some multithreaded computations, 8 CPUs and
as much as 16 GB of memory were used.

\subsection{Source code and software}
\label{sec:software}

The methods described in this paper were implemented in the {\tt
  fastTopics} R package. The main numerical results (other than the
  in-depth illustration with the MCF-7 data set) were generated using
  version 0.5-24 of the R package. The core optimization algorithms
  were developed in C++ and interfaced to R using {\tt
  Rcpp} \citep{rcpp}. The CD updates were adapted from the C++ code
  included with the NNLM R package, version 0.4-3 \citep{scd}.
%
% For the {\tt maptpx} results, we used a slightly
% modified version of the {\tt maptpx} package (version 1.9-8),
% available at \url{https://github.com/stephenslab/maptpx}, that allows
% initialization of both the topic proportions as well as the word
% frequencies.
%
% For the LDA results, we used the C implementation by Blei {\em et al}
% (\url{http://www.cs.columbia.edu/~blei/lda-c}), which was interfaced
% to R using the {\tt topicmodels} package, version 0.2-16
% \citep{topicmodels}. 
%
The Zenodo repository \citep{zenodo} (see also
\url{https://github.com/stephenslab/fastTopics-experiments/}) contains
the code implementing the numerical experiments, and a workflowr
website \citep{workflowr} for browsing the results.

\section{Additional figures}
\label{sec:extended-results}

\begin{figure*}[!ht]
\centering{
\includegraphics[width=\textwidth]{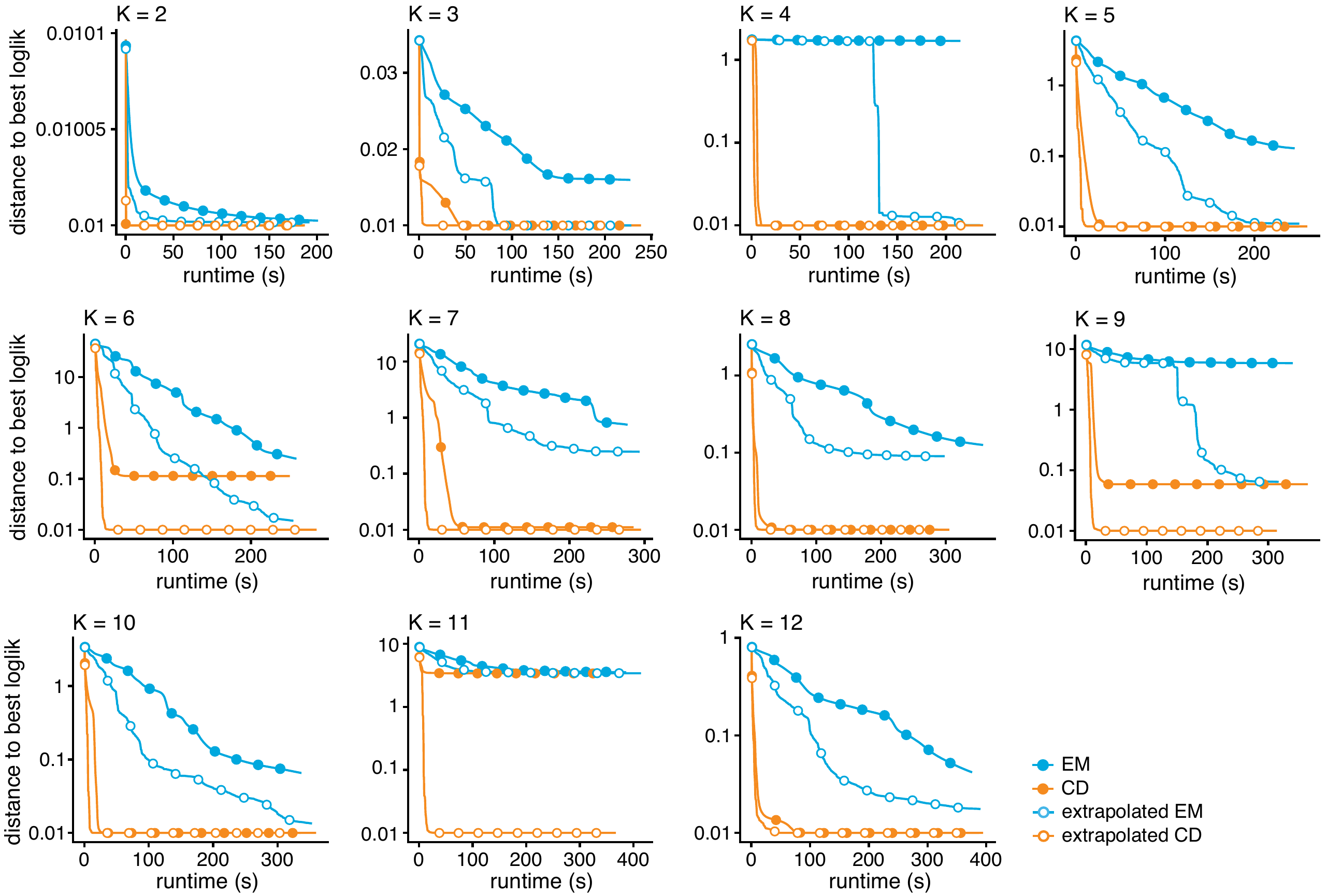}}
\caption{Improvement in model fit over time for the different Poisson
  NMF algorithms applied to the NeurIPS data. Multinomial topic model
  log-likelihoods are shown relative to the best log-likelihood
  recovered among the four algorithms compared (EM and CD, with and
  without extrapolation). Log-likelihood differences less than 0.01
  are shown as 0.01. Circles are drawn at intervals of 100
  iterations. Note that the 1,000 EM iterations performed during the
  initialization phase are not shown.}
\label{fig:loglik-nips}
\end{figure*}

\begin{figure*}[!t]
\centering{
\includegraphics[width=\textwidth]{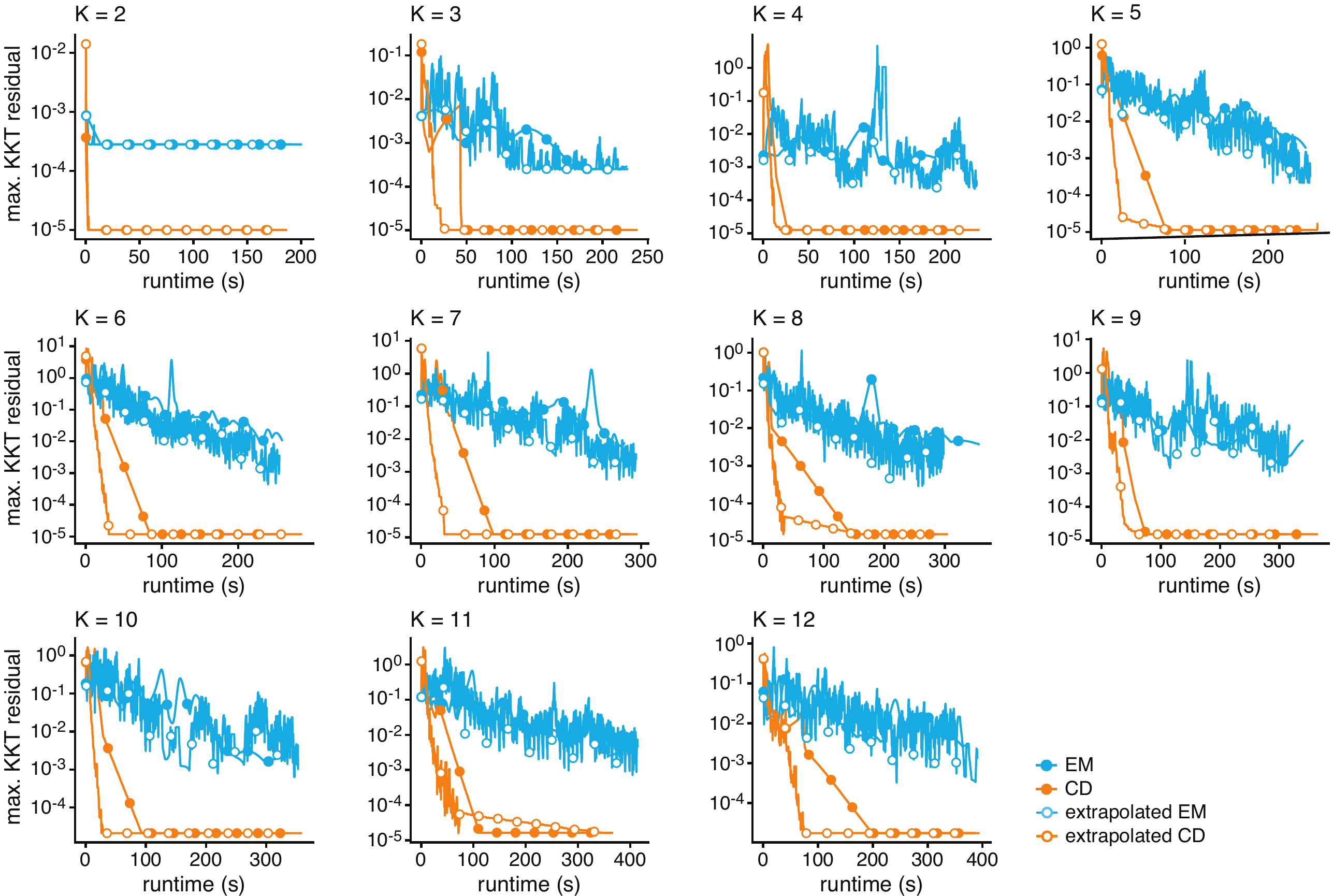}}
\caption{Evolution of the KKT residuals over time for the different
  Poisson NMF algorithms applied to the NeurIPS data. The KKT
  residuals should vanish near a local maximum of the Poisson NMF
  log-likelihood, so looking at the largest KKT residual can be used
  to assess how closely the algorithm recovers a stationary point
  ({\em i.e.}, an MLE). Note that the KKT residuals are not expected
  to decrease monotonically over time. Circles are drawn at intervals
  of 100 iterations.}
\label{fig:kkt-nips}
\end{figure*}

\begin{figure*}[!t]
\centering{
\includegraphics[width=\textwidth]{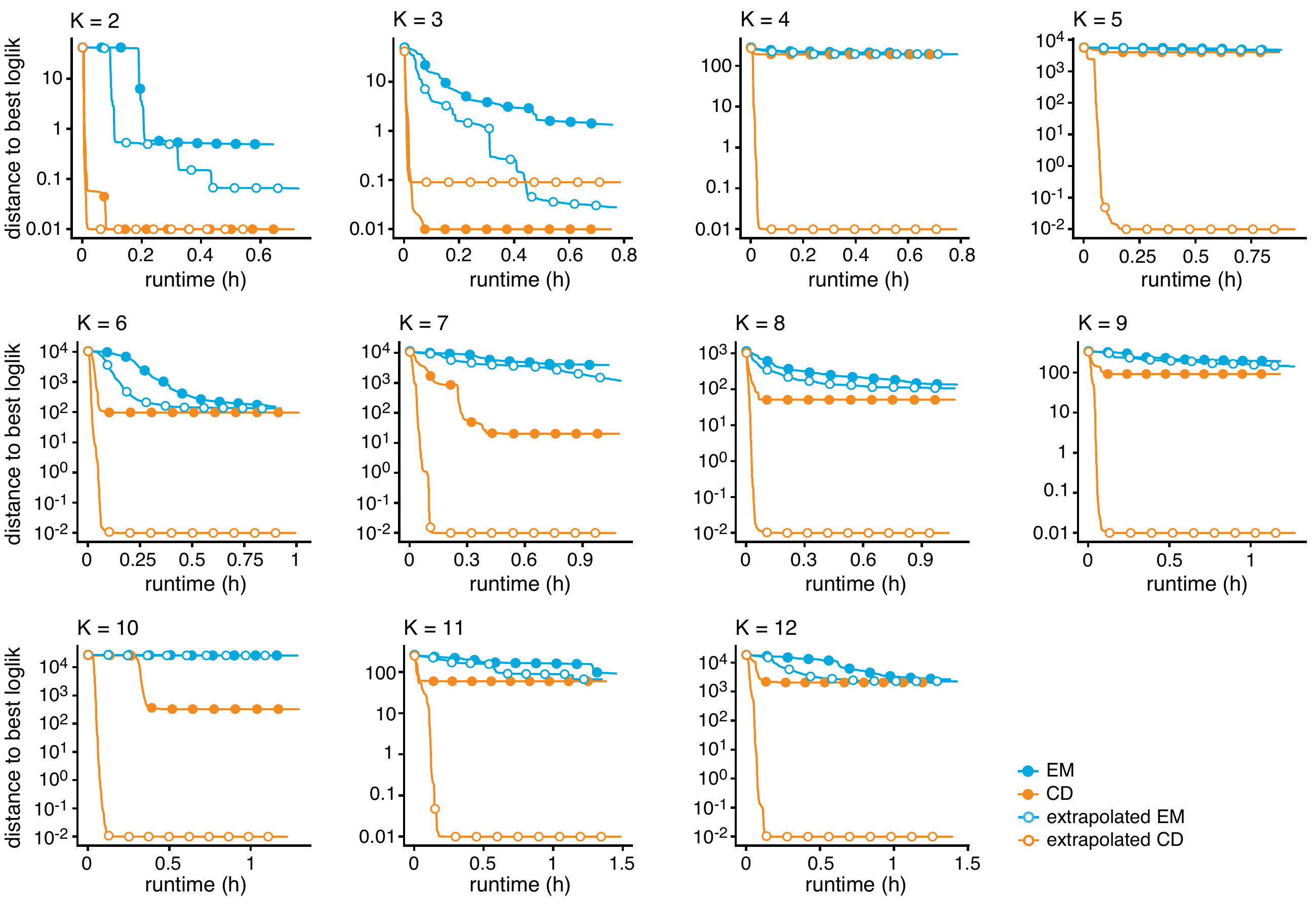}}
\caption{Improvement in model fit over time for the different Poisson
  NMF algorithms applied to the newsgroups data. See the
  Fig.~\ref{fig:loglik-nips} caption for more details.}
\label{fig:loglik-newsgroups}
\end{figure*}

\begin{figure*}[!t]
\centering{
\includegraphics[width=\textwidth]{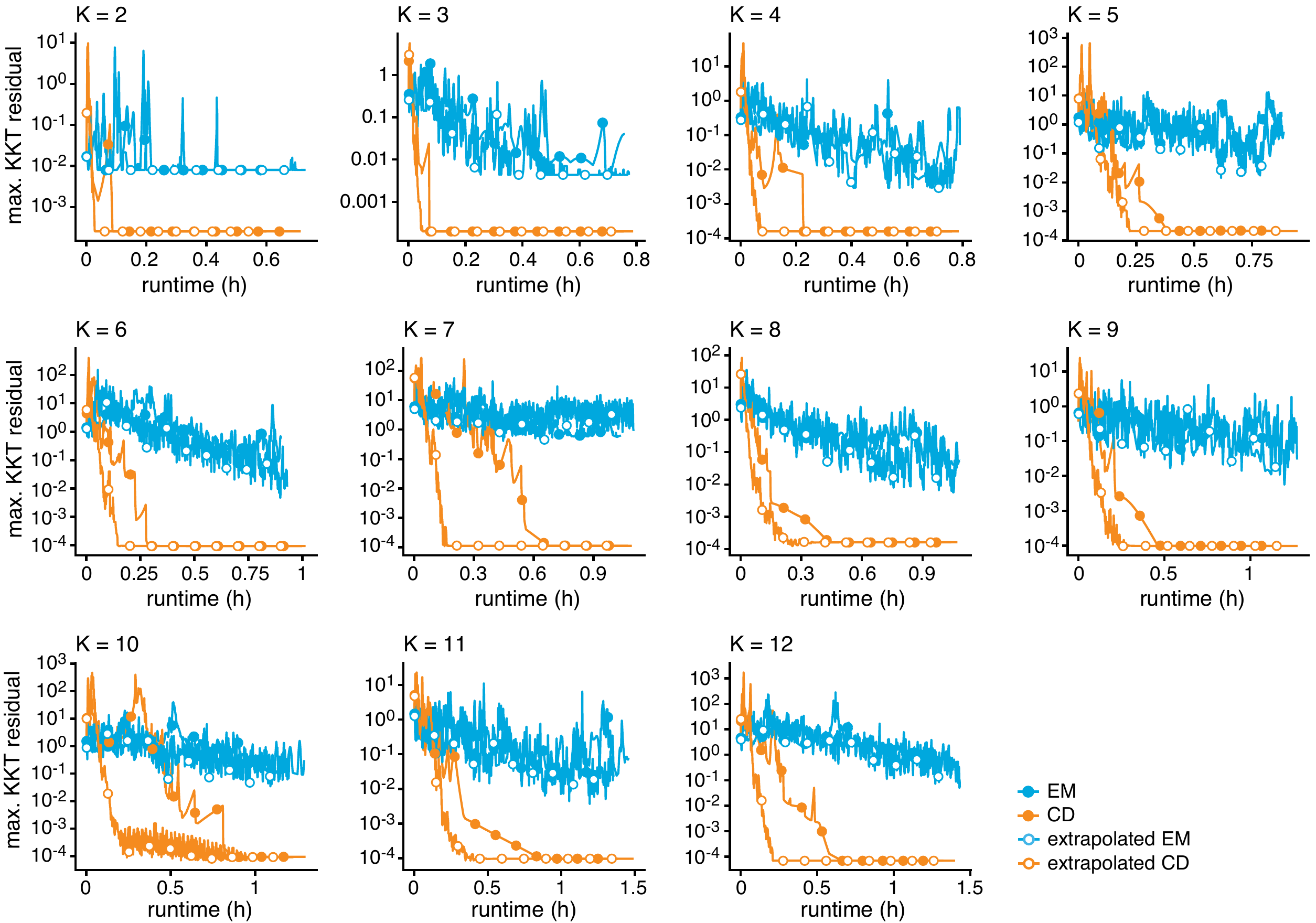}}
\caption{Evolution of the KKT residuals over time for the different
  Poisson NMF algorithms applied to the newsgroups data. See the
  Fig.~\ref{fig:kkt-nips} caption for more details.}
\label{fig:kkt-newsgroups}
\end{figure*}

\begin{figure*}[!t]
\centering{
\includegraphics[width=\textwidth]{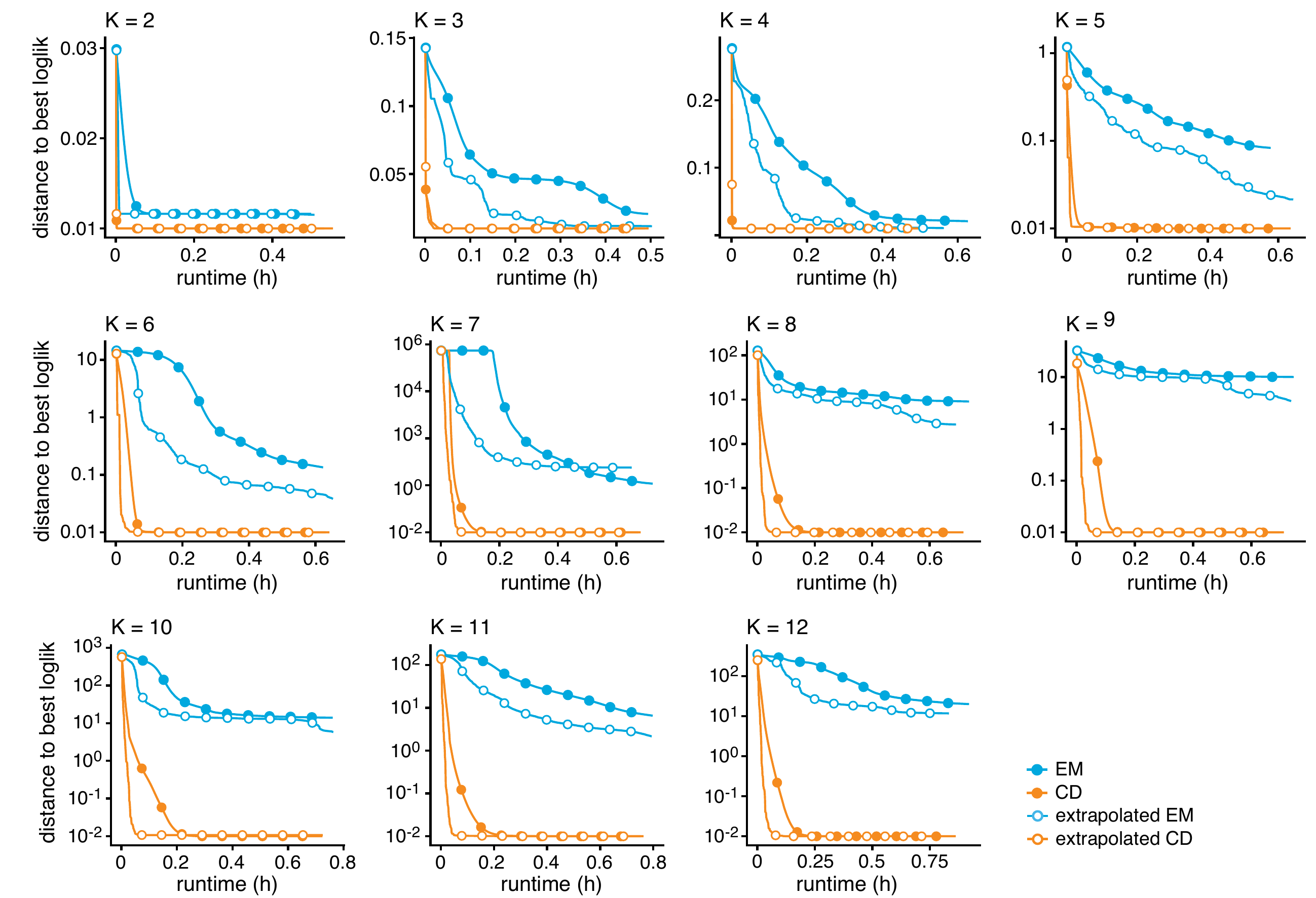}}
\caption{Improvement in model fit over time for the different Poisson
  NMF algorithms applied to the epithelial airway data. See the
  Fig.~\ref{fig:loglik-nips} caption for more details.}
\label{fig:loglik-droplet}
\end{figure*}

\begin{figure*}[!t]
\centering{
\includegraphics[width=\textwidth]{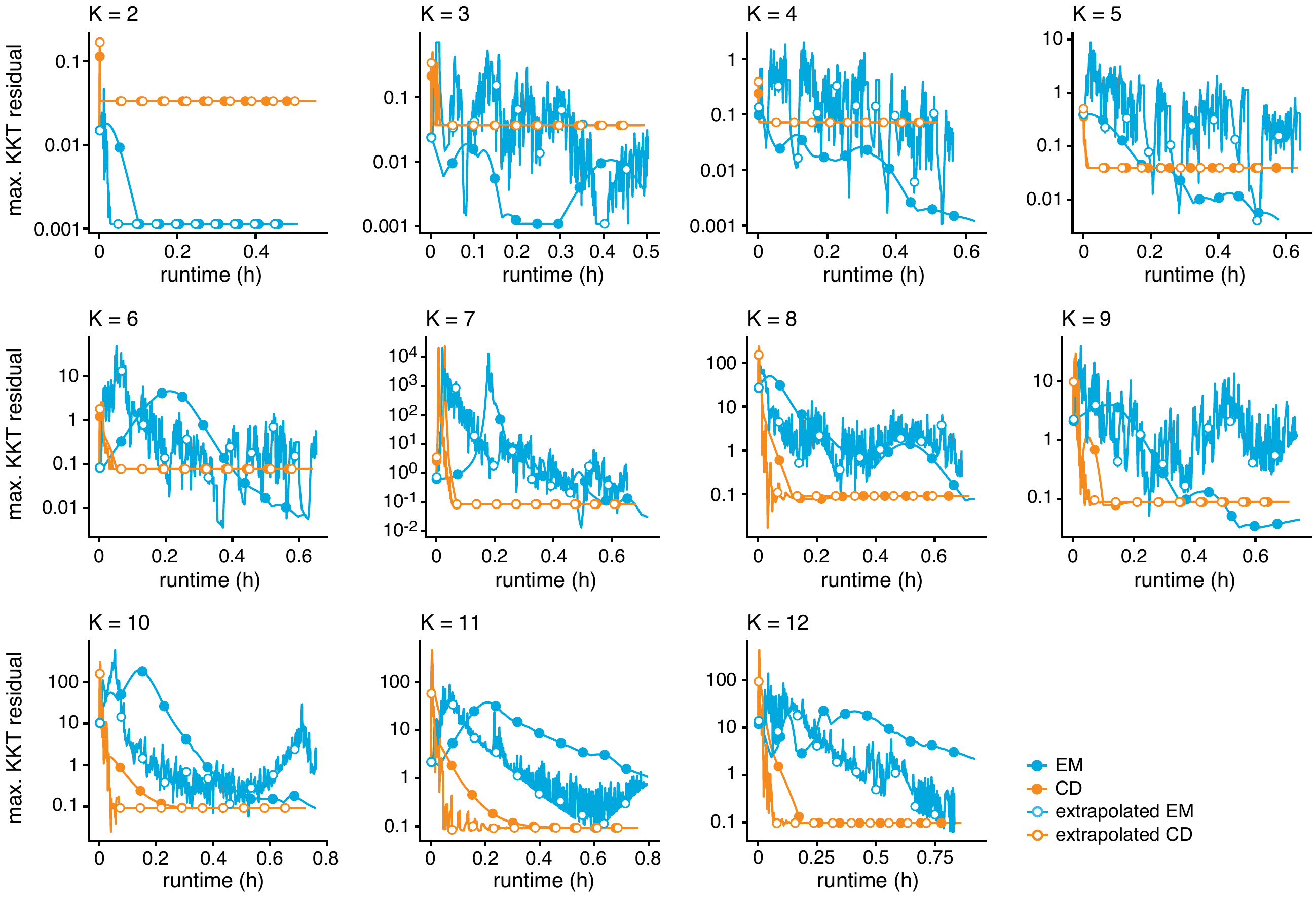}}
\caption{Evolution of the KKT residuals over time for the different
  Poisson NMF algorithms applied to the epithelial airway data. See
  the Fig.~\ref{fig:kkt-nips} caption for more details.}
\label{fig:kkt-droplet}
\end{figure*}

\begin{figure*}[!t]
\centering{
\includegraphics[width=\textwidth]{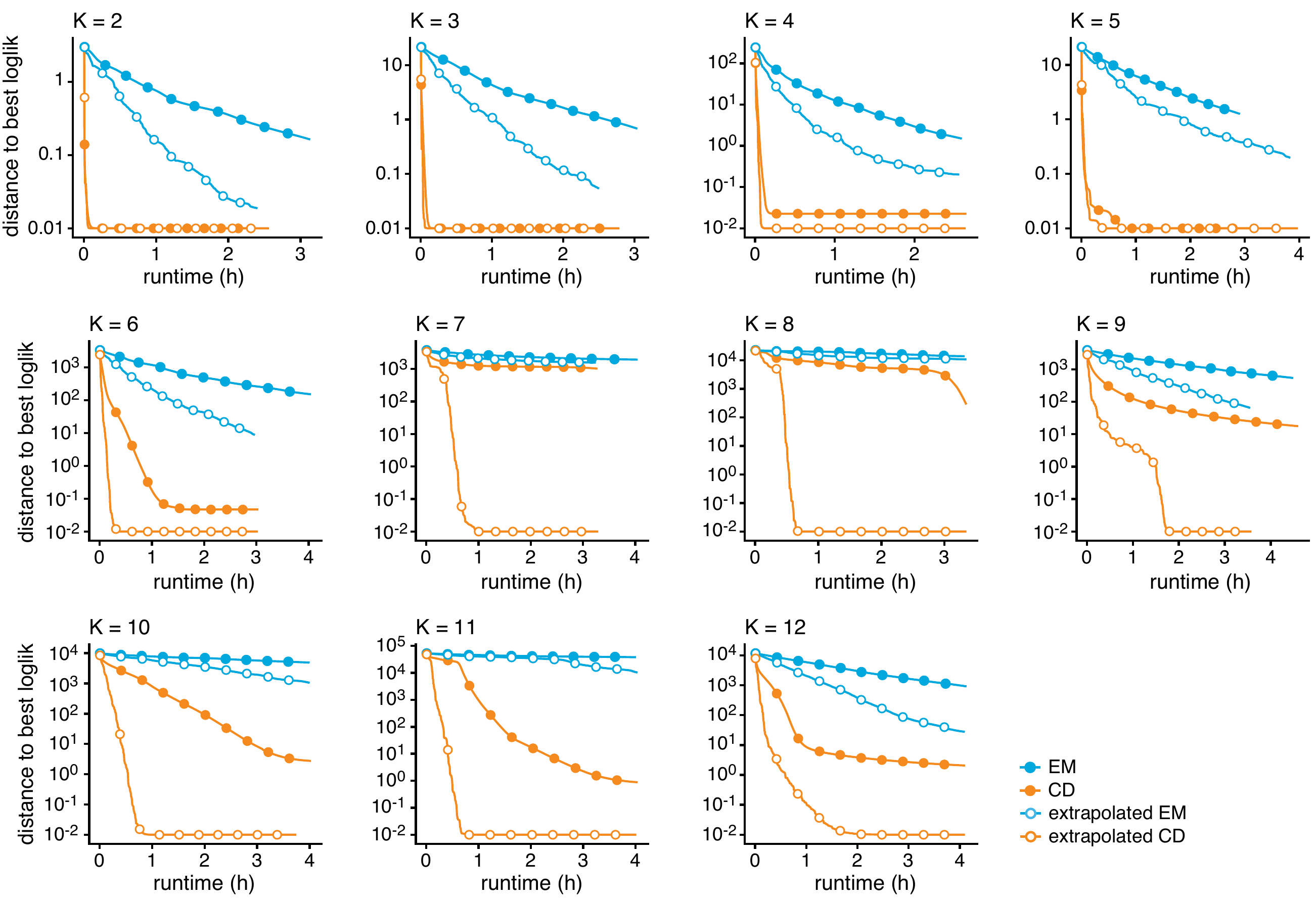}}
\caption{Improvement in model fit over time for the different Poisson
  NMF algorithms applied to the 68k PBMC data. See the
  Fig.~\ref{fig:loglik-nips} caption for more details.}
\label{fig:loglik-pbmc68k}
\end{figure*}

\begin{figure*}[!t]
\centering{
\includegraphics[width=\textwidth]{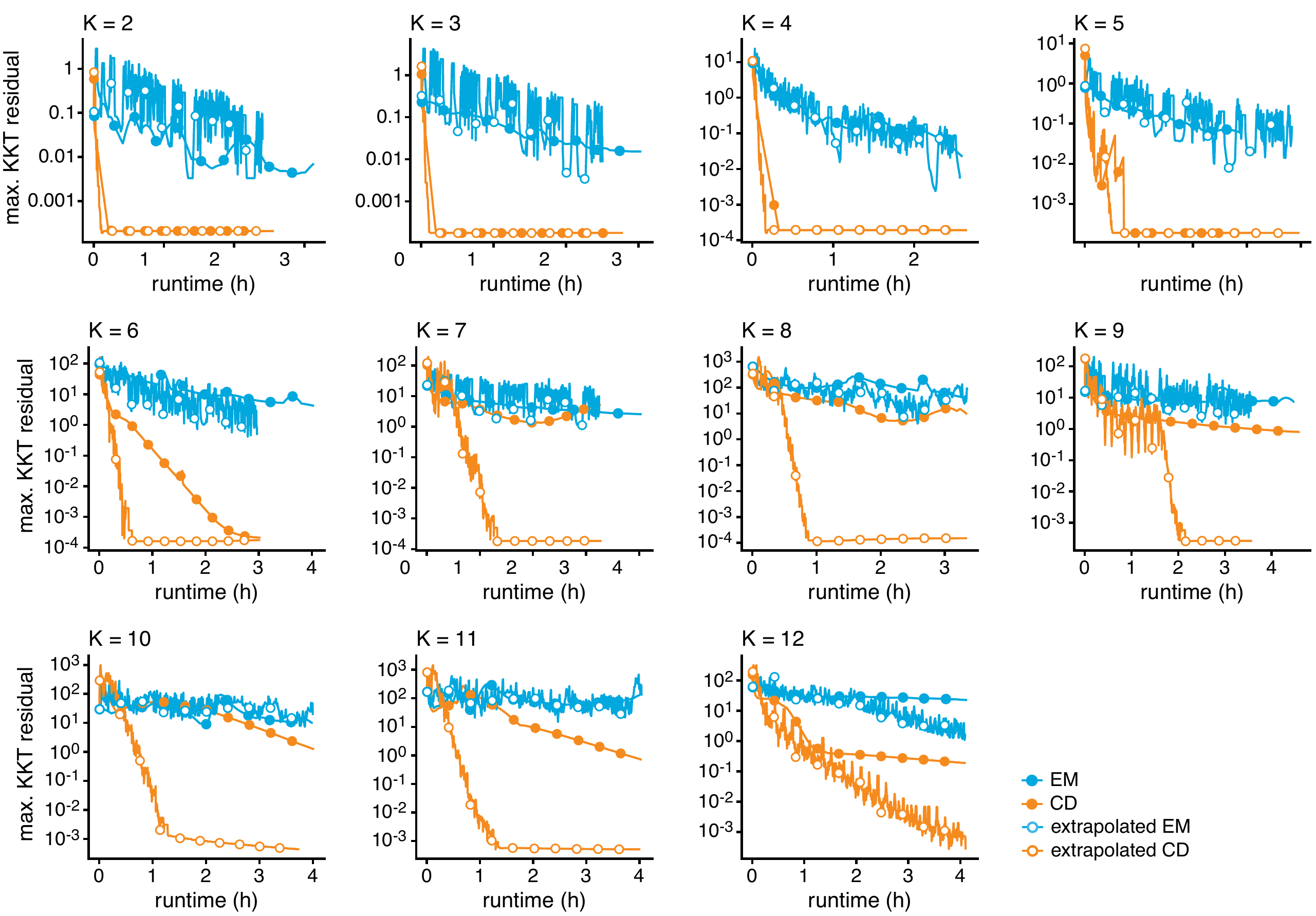}}
\caption{Evolution of the KKT residuals over time for the different
  Poisson NMF algorithms applied to the 68k PBMC data. See the
  Fig.~\ref{fig:kkt-nips} caption for more details.}
\label{fig:kkt-pbmc68k}
\end{figure*}

Figures \ref{fig:loglik-nips}--\ref{fig:kkt-pbmc68k} contain additional
results from the numerical experiments.

% Figure~\ref{fig:smallsim-maptpx} shows the results from running {\tt
%   maptpx} on the two simulated data sets (Sec.~\ref{sec:smallsim}).

% Figures \ref{fig:loglik-nips}--\ref{fig:kkt-pbmc68k} give more
% detailed results on the Poisson NMF algorithms' progress in fitting
% topic models to the real data sets (Sec.~\ref{sec:experiments}). The
% scatterplots in Figures \ref{fig:words-scatterplot-newsgroups} and
% \ref{fig:genes-scatterplot-pbmc68k} highlight words and genes that
% appear more frequently in one topic compared to the other topics.

% \begin{figure}[!t]
% \centering
% \includegraphics[width=0.75\textwidth]{smallsim_maptpx.pdf}
% \caption{{\em Maximum a posteriori} (MAP) estimation of the
%   multinomial topic model parameters, with $K = 6$, using the
%   quasi-Newton-accelerated EM algorithm implemented in {\tt maptpx}
%   \protect\citep{taddy-2012}. Plots A and B shows the
%   improvement in the log-posterior over time in Scenarios A and B, in
%   which the MAP estimates are initialized to MLEs obtained by
%   performing different numbers of EM and/or CD updates.}
% \label{fig:smallsim-maptpx}
% \end{figure}

% \begin{figure}[!t]
% \centering{
% \includegraphics[width=\textwidth]{elbo_nips.pdf}}
% \caption{Improvement in the ELBO over time for fitting LDA models to
%   the NeurIPS data. The LDA fits were initialized to the parameter
%   estimates obtained by running the four Poisson NMF algorithms (EM
%   and CD, with and without extrapolation). ELBOs are shown relative to
%   the best ELBO recovered among the four runs compared. ELBO
%   differences less than 1 are shown as 1.}
% \label{fig:elbo-nips}
% \end{figure}

% \begin{figure}[!t]
% \centering{
% \includegraphics[width=\textwidth]{elbo_newsgroups.pdf}}
% \caption{Improvement in the ELBO over time for fitting LDA models to
%   the newsgroups data. See the Figure~\ref{fig:elbo-nips} caption for
%   more details.}
% \label{fig:elbo-newsgroups}
% \end{figure}

% \begin{figure}[!t]
% \centering{
% \includegraphics[width=\textwidth]{elbo_droplet.pdf}}
% \caption{Improvement in the ELBO over time for fitting LDA models to
%   the epithelial airway data. See the Figure~\ref{fig:elbo-nips}
%   caption for more details.}
% \label{fig:elbo-droplet}
% \end{figure}

\end{appendices}

\backmatter

% \bmhead{Supplementary information}
% 
% If your article has accompanying supplementary file/s please state so here. 
% 
% Please refer to Journal-level guidance for any specific requirements.

\bmhead{Acknowledgements}

Many people have contributed helpful ideas and feedback, including
Mihai Anitescu, Kushal Dey, Adam Gruenbaum, Joyce Hsiao, Anthony Hung,
Youngseok Kim, Kaixuan Luo, John Novembre, Sebastian Pott, Alan
Selewa, Eric Weine and Jason Willwerscheid. We thank Xihui Lin, Paul
Boutros, Minzhe Wang and Tracy Ke for R code. And we thank the staff
at the Research Computing Center for providing the high-performance
computing resources used to implement the numerical experiments. This
work was supported by the NHGRI at the National Institutes of Health
under award number R01HG002585.

\bmhead{Author contributions}

P.C. and M.S. wrote the main manuscript. P.C. developed the software
and designed the experiments, with contributions from A.S., Z.W. and
M.S. All authors reviewed the manuscript.

\bmhead{Data availability}

All the data sets used in this paper are openly available: 
the NeurIPS data set was obtained from
\url{http://ai.stanford.edu/~gal/data.html}; 
the newsgroups data set was downloaded from
\url{http://qwone.com/~jason/20Newsgroups};
the MCF-7 and epithelial airway data sets were downloaded from the
Gene Expression Omnibus (GEO) website, accessions GSE152749 and
GSE103354; and the 68k PBMC data set was downloaded from the 
10x Genomics website,
\url{https://www.10xgenomics.com/datasets}.

\section*{Declarations}

\bmhead{Conflicts of interest}

The authors declare no conflicts of interest.


\setlength{\bibsep}{4pt}
\bibliography{fastTopics} % common bib file
% if required, the content of .bbl file can be included here once bbl
% is generated \input sn-article.bbl

\end{document}